# A Study on the Extraction and Analysis of a Large Set of Eye Movement Features during Reading

**Ioannis Rigas[a], Lee Friedman[a], Oleg Komogortsev[a]**

[a] **Department of Computer Science, Texas State University, San Marcos, USA**

rigas@txstate.edu, lfriedman10@gmail.com, ok@txstate.edu

## Abstract

This work presents a study on the extraction and analysis of a set of 101 categories of eye movement features from three types of eye movement events: fixations, saccades, and post-saccadic oscillations. The eye movements were recorded during a reading task. For the categories of features with multiple instances in a recording we extract corresponding feature subtypes by calculating descriptive statistics on the distributions of these instances. A unified framework of detailed descriptions and mathematical formulas are provided for the extraction of the feature set. The analysis of feature values is performed using a large database of eye movement recordings from a normative population of 298 subjects. We demonstrate the central tendency and overall variability of feature values over the experimental population, and more importantly, we quantify the test-retest reliability (repeatability) of each separate feature. The described methods and analysis can provide valuable tools in fields exploring the eye movements, such as in behavioral studies, attention and cognition research, medical research, biometric recognition, and human-computer interaction.

## 1. Introduction

The extraction of eye movement features that can model the characteristics of the oculomotor system's structure and functionality is a vital topic in many fields of research. Human eye movements are inherently connected to the guiding mechanisms of visual attention and thus can serve as an investigation tool for cognitive and behavioral studies. The classic study of (Yarbus, 1967) showed in a systematic way that the eye movements performed during the inspection of a visual stimulus are related to the performed cognitive task. With the advances in eye-tracking technology, analysis of eye movements was adopted as a useful tool for the observation of visual behavior in various studies of cognitive psychology in fields like linguistics, spatial processing, reading, and problem solving (Just & Carpenter, 1976; Rayner, 1998). Several research studies were also conducted to explore the underlying mechanisms connecting the generation of eye movements with visual attention and perception (Collins & Doré-Mazars, 2006; Eckstein, Beutter, Pham, Shimozaki, & Stone, 2007; Schütz, Braun, & Gegenfurtner, 2011). The increasing affordability of mobile eye-trackers facilitated the inspection of natural behavior in out-of-the-lab environments (Hayhoe & Ballard, 2005; Land, 2009).

The examination of eye movements can also facilitate studies focusing on the interconnections of the oculomotor behavior and individual characteristics. Eye movements have been explored in relation to individual motivation (Kaspar & König, 2011) and the 'Big 5' personality traits (agreeableness, conscientiousness, extraversion, neuroticism, and openness) (Rauthmann, Seubert, Sachse, & Furtner, 2012). Recently, vigor of eye movements was associated with the personal impulsiveness during decision-making tasks (Choi, Vaswani, & Shadmehr, 2014). Also, the research of personal eye movement traits has served as the basis for the field of eye movement biometrics (Rigas & Komogortsev, 2017).

The properties of eye movements have been investigated in medical research in order to indicate pathophysiological neural abnormalities, and for identifying early signs of neurodegenerative diseases (MacAskill & Anderson, 2016). For example, the reading paradigm was used in the past to explore the eye movement characteristics in cases of early Alzheimer disease (Fernández et al., 2013) and Parkinson disease (Wetzel, Gitchel, & Baron, 2011). The oculomotor behavior has been also studied in relation to various behavioral disorders such as ADHD (Fried et al., 2014) and autism (Klin, Jones, Schultz, Volkmar, & Cohen, 2002; Shirama, Kanai, Kato, & Kashino, 2016).

Given the large span of applications of eye movement analysis, the extraction of eye movement features has been always a fragmented research area, with most of the studies focusing on small sets of relevant each time features. This motivated our current work for the presentation of a unified framework for the extraction and analysis of a very large set of eye movement features. We focus on the reading paradigm since it allows for the exploration both of physical and of cognitive properties of the oculomotor activity. The extracted features cover a wide gamut of temporal, positional, and dynamic characteristics of three basic eye movement events: fixations, saccades and post-saccadic oscillations.

The contribution of our current work can be summarized as follows:

1) We present the largest to date collection of over 101 categories of eye movement features (and the deriving feature subtypes) and provide descriptions and formulas for their calculation from pre-classified eye movement events, specifically, fixations, saccades, and post-saccadic oscillations.
2) We use a large database from 298 subjects to calculate the values of central tendency (median) and overall variability (inter-quartile range) of the features extracted from our experimental paradigm.
3) We perform analysis of the test-retest reliability of the features by quantifying the absolute agreement of measurements using the Intraclass Correlation Coefficient (ICC) for normally distributed features and the Kendall’s coefficient of concordance (W) for non-normally distributed features.

## 2. Extraction of eye movement features

### *2.1. General overview of feature extraction and used notation*

Prior to feature extraction, the raw eye movement recordings need to be preprocessed in order to detect (classify) the signal parts that correspond to the basic types of eye movement events, namely fixations saccades, and post-saccadic oscillations (see definitions in next sections). The algorithm used to perform this classification was a modified version of the velocity-based method presented in (Nyström & Holmqvist, 2010). The performed modifications focused on the adoption of specific values for the thresholds that would allow the optimum classification performance for the case of data recorded for the experimental paradigm of reading text. The accuracy of the classification results was additionally verified via visual inspection of the classified eye movement events.

In general, the extracted features can be categorized in two basic groups: **non-distributional features** and **distributional features**. As non-distributional features we denote the features for which a single value is calculated for each recording by building a collective model over all instances of an event (fixation, saccade or post-saccadic oscillation). As distributional features we denote the features for which a separate value is extracted for every classified instance of an event, and then, six different descriptive statistics are applied on the distribution of these multiple values to generate six corresponding feature subtypes. The used statistics are the mean (*MN*), median (*MD*), standard deviation (*SD*), interquartile range (*IQ*), skewness (*SK*), and kurtosis (*KU*).

In List 2.1.1 we present various symbols and notation used in the descriptions that follow.

**List 2.1.1** Symbols and notation

$\boldsymbol{Fix^{Num}}, \boldsymbol{Sac^{Num}}, \boldsymbol{Gls^{Num}}$: denote the total number of fixations, saccades, and post-saccadic oscillations in a recording

$\boldsymbol{FixPos_i(j)}, \boldsymbol{SacPos_i(j)}, \boldsymbol{GlsPos_i(j)}$: denote the $j^{th}$ positional sample of $i^{th}$ event (fixation, saccade, or post-saccadic oscillation)

$\boldsymbol{FixVel_i(j)}, \boldsymbol{SacVel_i(j)}, \boldsymbol{GlsVel_i(j)}$: denote the $j^{th}$ velocity sample of $i^{th}$ event (fixation, saccade, or post-saccadic oscillation)

$\boldsymbol{FixAcc_i(j)}, \boldsymbol{SacAcc_i(j)}, \boldsymbol{GlsAcc_i(j)}$: denote the $j^{th}$ acceleration sample of $i^{th}$ event (fixation, saccade, or post-saccadic oscillation)

$\boldsymbol{DistrStat}$: used as superscript to denote distributional features. Feature subtypes from feature instances $x_i$ are generated as $DistrStat(x_i)$, where $DistrStat(\cdot)$ = *mean, median, standard deviation, interquartile range, skewness, and kurtosis*

$\boldsymbol{HVR}$ *or* $\boldsymbol{HV}$ *or* $\boldsymbol{R}$: used as superscript to denote features calculated for horizontal-vertical-radial or horizontal-vertical-only or radial-only components of eye movement.

$\boldsymbol{HV2D}$: used as superscript to denote features calculated to represent position samples in 2-D plane (2D-trajectory)

## *2.2. Fixation features*

The term fixation is used to define the state when the eyes are focused on a specific area of interest, projecting the content of this area on the high-resolution processing region of the retina (fovea centralis). During fixation the eyes are not totally still but they perform various miniature movements: slow ocular drifts, small saccades (micro-saccades), and high-frequency tremors (sometimes referred as physiological nystagmus) (Steinman, Haddad, A.A., & Wyman, 1973). In next sections we present the features that can be extracted from fixations, grouped according to the eye movement characteristics that they model, i.e. temporal, positional, or dynamic characteristics.

### *2.2.1. Features of fixation temporal characteristics*

The fixation duration and fixation rate (*F01-F02*, List 2.2.1) are two very basic features that can provide valuable information about the temporal behavior of the oculomotor functionality. In the case of the reading paradigm, the temporal characteristics of fixations can be used to examine the cognitive function that can be co-modulated by various aspects such as the context (Raney, Campbell, & Bovee, 2014), subject-related idiosyncrasies (Holland & Komogortsev, 2011), and the mental workload (Ahram et al., 2015). In Fig. 1 (left), we show a sequence of fixations performed during reading where we can observe the existing variability in durations of the performed fixations.

**List 2.2.1** Fixation temporal features

$\boldsymbol{F01}$: $\boldsymbol{Fix_{Rate}}$
The fixation rate: $Fix^{Num}/Rec^{dur}$ , $Rec^{dur}$: total recording duration

$\boldsymbol{F02}$: $\boldsymbol{FixDur^{DistrStat}}$
$DistrStat(\cdot)$ on durations of fixations: $FixDur_i, i = 1, \ldots, Fix^{Num}$

### *2.2.2. Features of fixation position and drift*

A simple way to model the position of a fixation is to calculate the centroid (*F03,* List 2.2.2) over all samples within the position profile*. This simple feature, though, cannot model the characteristics of fixation drift, i.e. the slow movement of the eye around a fixated location. The computational modeling of fixation drift can provide information about the stability of the visual input in the retina and related cognitive implications (Poletti, Listorti, & Rucci, 2010), and also, it can be used as a cue for the detection of pathological conditions like amblyopia (Schor & Westall, 1984) and cerebellar disease (Leech, Gresty, Hess, & Rudge, 1977). It should be mentioned that the fixation drift can be also attributed to device dependent sources (sometimes called ‘baseline drift’), and so, the modeling of fixation drift can be particularly useful for human-computer interaction applications (Stampe & Reingold, 1995) and during the inspection of eye-tracking quality (Hornof & Halverson, 2002). The fixation drift can be manifested in a variety of forms (see Fig. 1, right) and for this reason we present a number of alternative features that can be extracted to model the characteristics of fixation drift (*F04-F13*, List 2.2.2).

* The term *profile* refers to the variation of a quantity (position, velocity, acceleration) as a function of time/samples

**List 2.2.2** Fixation position and drift features

$\boldsymbol{F03}$: $\boldsymbol{FixPosCentroid^{DistrStat-HV}}$
$DistrStat(\cdot)$ on position centroids of fixations: $FixPosCentroid_i = \sum_{j=1}^{N} FixPos_i(j)/N\,,\ i = 1, \ldots, Fix^{Num}$

$\boldsymbol{F04}$: $\boldsymbol{FixDriftDisp^{DistrStat-HVR}}$
$DistrStat(\cdot)$ on drift displacements of fixations: $FixDriftDisp_i = |FixPos_i(end) - FixPos_i(start)|, i = 1, \ldots, Fix^{Num}$

$\boldsymbol{F05}$: $\boldsymbol{FixDriftDist^{DistrStat-HVR}}$
$DistrStat(\cdot)$ on drift distances of fixations: $FixDriftDist_i = \sum_{j=1}^{N-1}|FixPos_i(j+1) - FixPos_i(j)|,\ i = 1, \ldots, Fix^{Num}$

$\boldsymbol{F06}$: $\boldsymbol{FixDriftAvgSpeed^{DistrStat-HVR}}$
$DistrStat(\cdot)$ on drift average speeds of fixations: $FixDriftAvgSpeed_i = FixDriftDisp_i/FixDur_i, i = 1, \ldots, Fix^{Num}$

$\boldsymbol{F07}$: $\boldsymbol{FixDriftFitLn_{Slope}^{DistrStat-HV}}$
$DistrStat(\cdot)$ on drift linear-regression-fit slope of fixations: $FixDriftFitLn_{Slope_i}$ calculated via linear regression fit on positional samples $FixPos_i(j),\ j = 1, \ldots, N$ within each fixation $i, i = 1, \ldots, Fix^{Num}$

$\boldsymbol{F08}$: $\boldsymbol{FixDriftFitLn_{R^2}^{DistrStat-HV}}$
$DistrStat(\cdot)$ on drift linear-regression-fit $R^2$ of fixations: $FixDriftFitLn_{R^2{}_i}$ calculated via linear regression fit on positional samples $FixPos_i(j),\ j = 1, \ldots, N$ within each fixation $i, i = 1, \ldots, Fix^{Num}$

$\boldsymbol{F09}$: $\boldsymbol{FixDriftFitQd_{R^2}^{DistrStat-HV}}$
$DistrStat(\cdot)$ on drift quadratic-regression-fit $R^2$ of fixations: $FixDriftFitQd_{R^2{}_i}$ calculated via quadratic regression fit on positional samples $FixPos_i(j),\ j = 1, \ldots, N$ within each fixation $i, i = 1, \ldots, Fix^{Num}$

$\boldsymbol{F10}$: $\boldsymbol{FixDriftPrL0Q0^{HV}}$
The L0Q0 parameter* percentage: $100\% \cdot \sum_{i=1}^{Fix^{Num}} L0Q0_i/Fix^{Num}$, with the $L0Q0_i$ calculated via stepwise multilinear regression fit on all positional samples $FixPos_i(j), j = 1, \ldots, N$ within each fixation $i, i = 1, \ldots, Fix^{Num}$

$\boldsymbol{F11}$: $\boldsymbol{FixDriftPrL0Q1^{HV}}$
The L0Q1 parameter percentage: $100\% \cdot \sum_{i=1}^{Fix^{Num}} L0Q1_i/Fix^{Num}$, with the $L0Q1_i$ calculated via stepwise multilinear regression fit on all positional samples $FixPos_i(j), j = 1, \ldots, N$ within each fixation $i, i = 1, \ldots, Fix^{Num}$

$\boldsymbol{F12}$: $\boldsymbol{FixDriftPrL1Q0^{HV}}$
The L1Q0 parameter percentage: $100\% \cdot \sum_{i=1}^{Fix^{Num}} L1Q0_i/Fix^{Num}$, with the $L1Q0_i$ calculated via stepwise multilinear regression fit on all positional samples $FixPos_i(j), j = 1, \ldots, N$ within each fixation $i, i = 1, \ldots, Fix^{Num}$

$\boldsymbol{F13}$: $\boldsymbol{FixDriftPrL1Q1^{HV}}$
The L1Q1 parameter percentage: $100\% \cdot \sum_{i=1}^{Fix^{Num}} L1Q1_i/Fix^{Num}$, with the $L1Q1_i$ calculated via stepwise multilinear regression fit on all positional samples $FixPos_i(j), j = 1, \ldots, N$ within each fixation $i, i = 1, \ldots, Fix^{Num}$

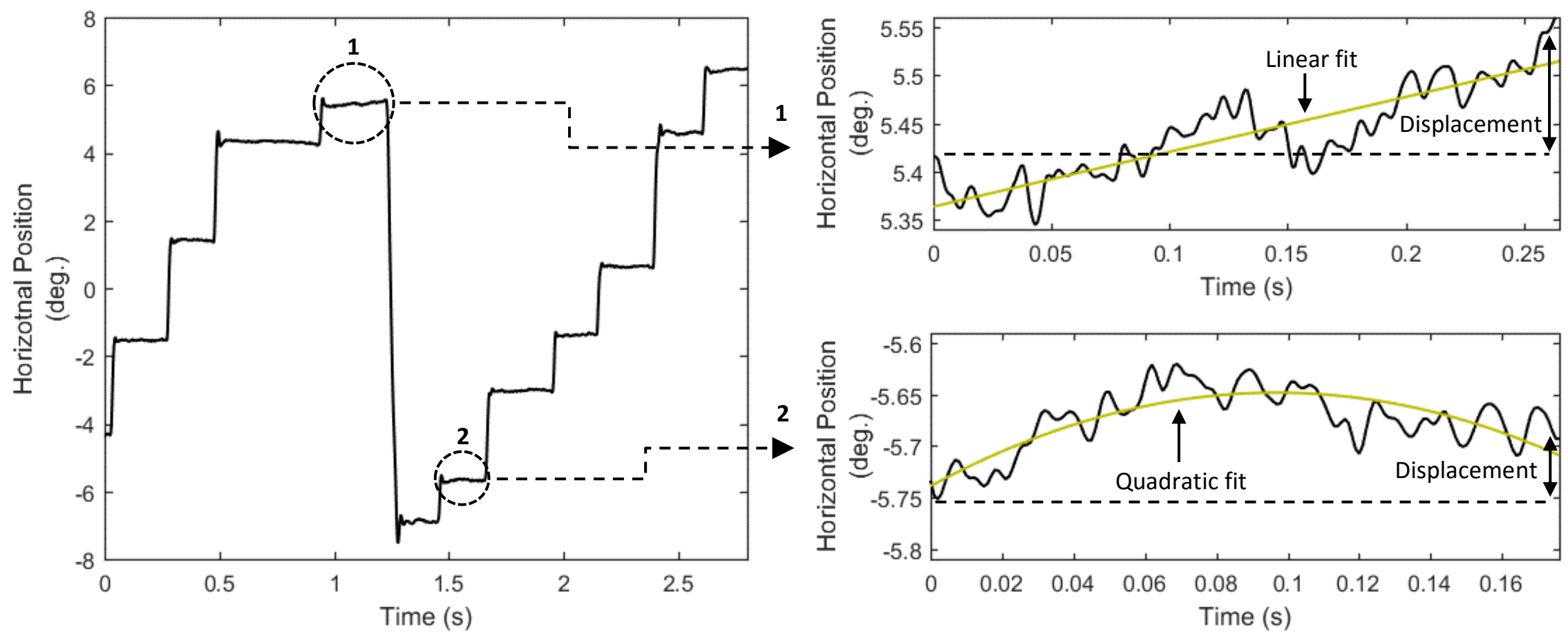

**Fig. 1** Eye movement positional signal and two examples of different modeling of fixation drifts (right top-linear fit is preferred; right bottom-quadratic fit is preferred).

* LxQy parameters show whether linear or/and quadratic terms were used for regression, e.g. $L0Q1_i = 1$ when only quadratic terms where used, $L1Q0_i = 1$ when only linear terms where used, etc.

*2.2.3. Features of fixation velocity and acceleration*

During fixation there is limited eye mobility and thus the fixation velocity and acceleration profiles are usually affected by noise. However, the variability in these profiles can also reflect information stemming from physiological sources, e.g. micro-movements and oculomotor function irregularities. Various studies have examined these movements in connection to visual perception (Martinez-Conde, Macknik, & Hubel, 2004) or for the examination of pathological conditions (Bolger et al., 2000; Bylsma et al., 1995). In Lists 2.2.3.1-2.2.3.2 we present the features that model fixation velocity and acceleration (*F14-F25*), and in Fig. 2-3 we show examples of fixation velocity and acceleration profiles. Most features are extracted via the statistical modeling of profiles using five descriptive statistics: mean, median, standard deviation, skewness and kurtosis. Such profile-modeling features have been previously employed in eye movement biometrics (George & Routray, 2016) both for fixations and saccades. It is important to clarify that the statistical modeling of profiles should not be confused with the mechanism used for the creation of feature subtypes (described in Section 2.1). The employed statistics are similar but in this case we use them to model the profiles of different instances of events.

**List 2.2.3.1** Fixation velocity features

$\boldsymbol{F14}$: $\boldsymbol{FixVelProfMn^{DistrStat-HVR}}$
$DistrStat(\cdot)$ on velocity profile-sample mean of fixations: $FixVelProfMn_i = \sum_{j=1}^{N}|FixVel_i(j)|/N\,, i = 1, \ldots, Fix^{Num}$

$\boldsymbol{F15}$: $\boldsymbol{FixVelProfMd^{DistrStat-HVR}}$
$DistrStat(\cdot)$ on velocity profile-sample median of fixations: $FixVelProfMd_i = median(|FixVel_i|), i = 1, \ldots, Fix^{Num}$

$\boldsymbol{F16}$: $\boldsymbol{FixVelProfSd^{DistrStat-HVR}}$
$DistrStat(\cdot)$ on velocity profile-sample standard deviation of fixations: $FixVelProfSd_i = \sqrt{\sum_{j=1}^{N}(|FixVel_i(j)| - FixVelProfMn_i)^2/N}\,, i = 1, \ldots, Fix^{Num}$

$\boldsymbol{F17}$: $\boldsymbol{FixVelProfSk^{DistrStat-HVR}}$
$DistrStat(\cdot)$ on velocity profile-sample skewness of fixations: $FixVelProfSk_i = \frac{\sum_{j=1}^{N}(|FixVel_i(j)|-FixVelProfMn_i)^3/N}{\left(\sqrt{\sum_{j=1}^{N}(|FixVel_i(j)|-FixVelProfMn_i)^2/N}\right)^3}, i = 1, \ldots, Fix^{Num}$

$\boldsymbol{F18}$: $\boldsymbol{FixVelProfKu^{DistrStat-HVR}}$
$DistrStat(\cdot)$ on velocity profile-sample kurtosis of fixations: $FixVelProfKu_i = \frac{\sum_{j=1}^{N}(|FixVel_i(j)|-FixVelProfMn_i)^4/N}{\left(\sum_{j=1}^{N}(|FixVel_i(j)|-FixVelProfMn_i)^2/N\right)^2}, i = 1, \ldots, Fix^{Num}$

$\boldsymbol{F19}$: $\boldsymbol{FixPrAbVelP90^{DistrStat-R}}$
$DistrStat(\cdot)$ on percentages of the velocity samples of fixations that are above $FixVelP90$ (90-th percentile) threshold: $FixPrAbVelP90_i, i = 1, \ldots, Fix^{Num}$

$\boldsymbol{F20}$: $\boldsymbol{FixPrCrVelP90^{DistrStat-R}}$
$DistrStat(\cdot)$ on percentages of the velocity samples of fixations that cross $FixVelP90$ (90-th percentile) threshold: $FixPrCrVelP90_i, i = 1, \ldots, Fix^{Num}$

**List 2.2.3.2** Fixation acceleration features

$\boldsymbol{F21}$: $\boldsymbol{FixAccProfMn^{DistrStat-HVR}}$
$DistrStat(\cdot)$ on acceleration profile-sample mean of fixations: $FixAccProfMn_i = \sum_{j=1}^{N} |FixAcc_i(j)|/N\ , i = 1, \ldots, Fix^{Num}$

$\boldsymbol{F22}$: $\boldsymbol{FixAccProfMd^{DistrStat-HVR}}$
$DistrStat(\cdot)$ on acceleration profile-sample median of fixations: $FixAccProfMd_i = median(|FixAcc_i|), i = 1, \ldots, Fix^{Num}$

$\boldsymbol{F23}$: $\boldsymbol{FixAccProfSd^{DistrStat-HVR}}$
$DistrStat(\cdot)$ over acceleration profile-sample standard deviation of fixations: $FixAccProfSd_i = \sqrt{\sum_{j=1}^{N} (|FixAcc_i(j)| - FixAccProfMn_i)^2/N}\ , i = 1, \ldots, Fix^{Num}$

$\boldsymbol{F24}$: $\boldsymbol{FixAccProfSk^{DistrStat-HVR}}$
$DistrStat(\cdot)$ on acceleration profile-sample skewness of fixations: $FixAccProfSk_i = \frac{\sum_{j=1}^{N} (|FixAcc_i(j)| - FixAccProfMn_i)^3/N}{\left(\sqrt{\sum_{j=1}^{N} (|FixAcc_i(j)| - FixAccProfMn_i)^2/N}\right)^3}, i = 1, \ldots, Fix^{Num}$

$\boldsymbol{F25}$: $\boldsymbol{FixAccProfKu^{DistrStat-HVR}}$
$DistrStat(\cdot)$ on acceleration profile-sample kurtosis of fixations: $FixAccProfKu_i = \frac{\sum_{j=1}^{N} (|FixAcc_i(j)| - FixAccProfMn_i)^4/N}{\left(\sum_{j=1}^{N} (|FixAcc_i(j)| - FixAccProfMn_i)^2/N\right)^2}, i = 1, \ldots, Fix^{Num}$

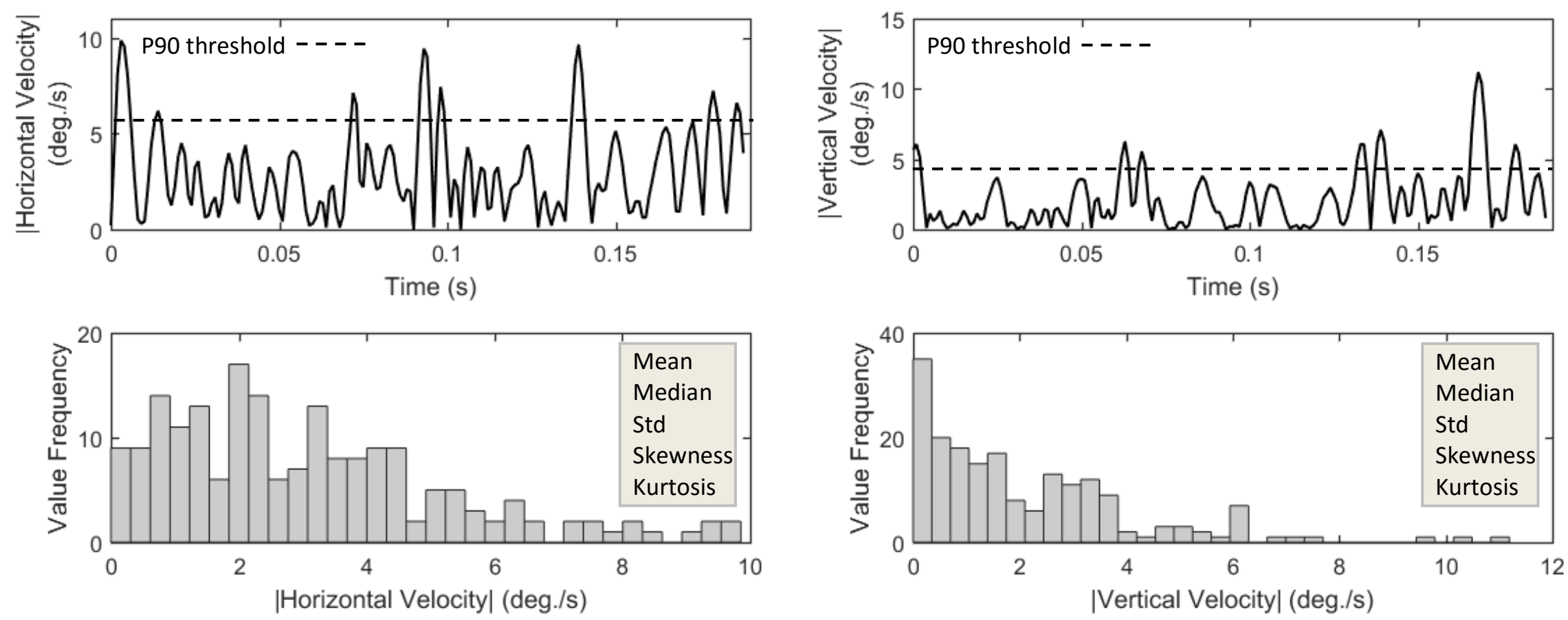

**Fig. 2** Horizontal and vertical fixation velocity profiles (top) and the respective distributions (bottom).

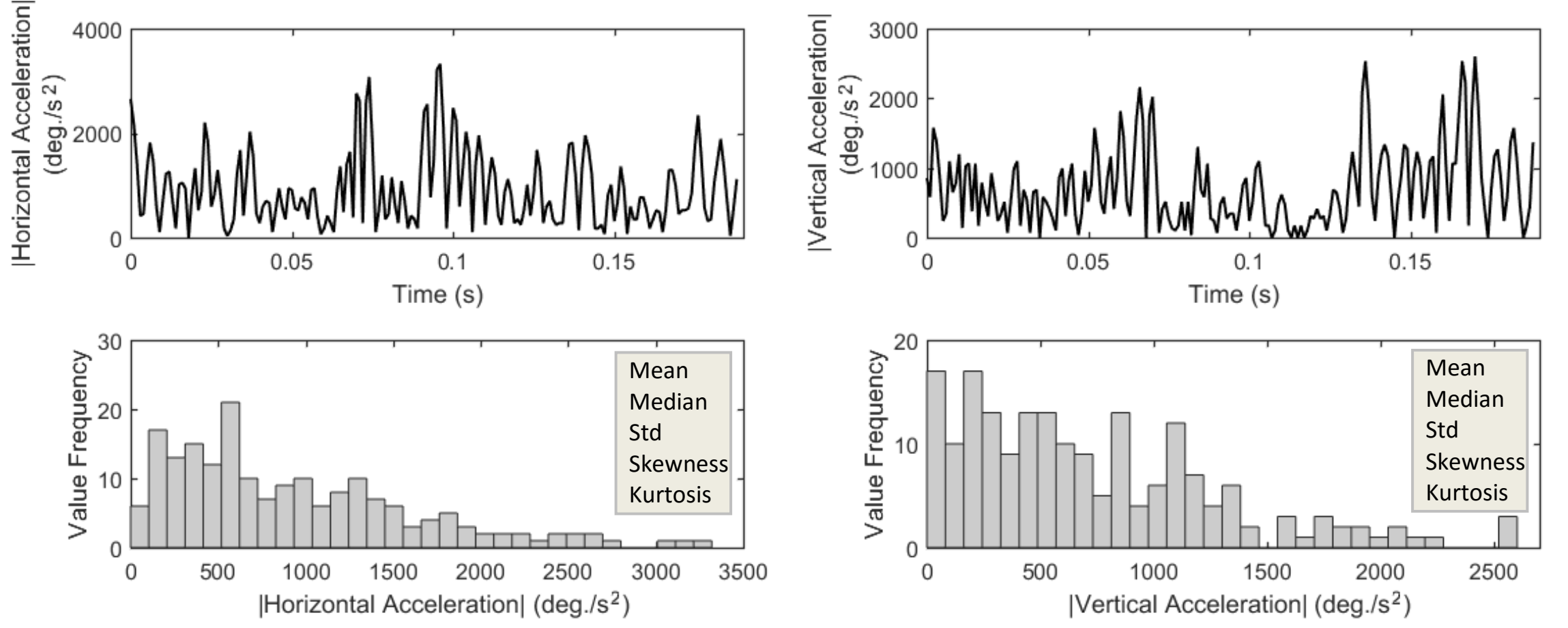

**Fig. 3** Horizontal and vertical fixation acceleration profiles (top) and the respective distributions (bottom).

### *2.3. Saccade features*

The saccades are very fast movements that rotate the eyes from one position of focus to another with peak velocities that can reach over 600°/s. For the initiation of a saccade, the saccade-generating neural circuitry estimates the difference between the starting and target position and sends neural guiding pulses to the extra-ocular muscles that rotate the eye. If the intended target is not accurately reached one or more small corrective saccades are performed to transfer the eye to target position. For the calculation of saccade features, we perform post-processing of data for filtering out any saccades with durations larger than 70 ms and radial sizes larger than 8° (and the adjacent post-saccadic oscillations). Such instances are expected to generate outliers to normal distributions of saccade features, given that during reading usually occur relatively small saccades. An exception was made for features *S49*-*S52*, since the role of these features is exactly to measure the frequency of these large saccadic events.

#### *2.3.1. Features of saccade temporal characteristics*

The two basic temporal features of saccades are duration and rate (*S01*-*S02*, List 2.3.1). For reading stimulus, saccade durations usually lie in range of 20-40 ms. The incorporation (or not) of saccadic durations during the analysis of eye movement data is an important topic in studies of cognitive processing (Inhoff & Radach, 1998) and it has been also explored for tasks of perceptual selection in human-computer interaction (Canosa, 2009). Atypical values of saccade temporal features (e.g. larger than usual durations) can signal the onset of neural disorders (Ramat, Leigh, Zee, & Optican, 2007). Also, the appearance of increased saccadic rates has been reported in studies of behavioral disorders like autism (Kemner, Verbaten, Cuperus, Camfferman, & van Engeland, 1998).

**List 2.3.1** Saccade temporal features

$\boldsymbol{S01}: \boldsymbol{Sac_{Rate}}$
The saccade rate: $Sac^{Num}/Rec^{dur}$

$\boldsymbol{S02}: \boldsymbol{SacDur^{DistrStat}}$
$DistrStat(\cdot)$ on durations of saccades: $SacDur_i, i = 1, \dots, Sac^{Num}$

#### *2.3.2. Features of saccade amplitude and curvature*

The amplitude of a saccade (*S03*, List 2.3.2) is frequently used as the basic feature to describe its size. In general, saccade amplitudes are linearly related to the respective durations and approximately logarithmically related to respective peak velocities (Bahill, Clark, & Stark, 1975). However, the saccade amplitude cannot be used to describe the curved trajectory usually exhibited by saccades. The modeling of saccade curvature can be very useful for behavioral studies since this characteristic has been connected to the distractor-related modulation of eye movements (Doyle & Walker, 2001). The representation of saccade curvature has been thoroughly reviewed in (Ludwig & Gilchrist, 2002), where a large variety of curvature features (old and new) were described. We have included these features in the current set (*S08*-*S19*, List 2.3.2) along with additional features that model non-linearity of saccadic trajectories (*S04*-*S05*, List 2.3.2). We observed that the ending parts of saccades often show larger degree of non-linearity, and thus, we present two more features for modeling the saccade ending part (‘tail’) (*S06*-*S07*, List 2.3.2). In Fig.4, we show examples of saccade trajectories both in position-time domain (position profile) and in 2D-domain.

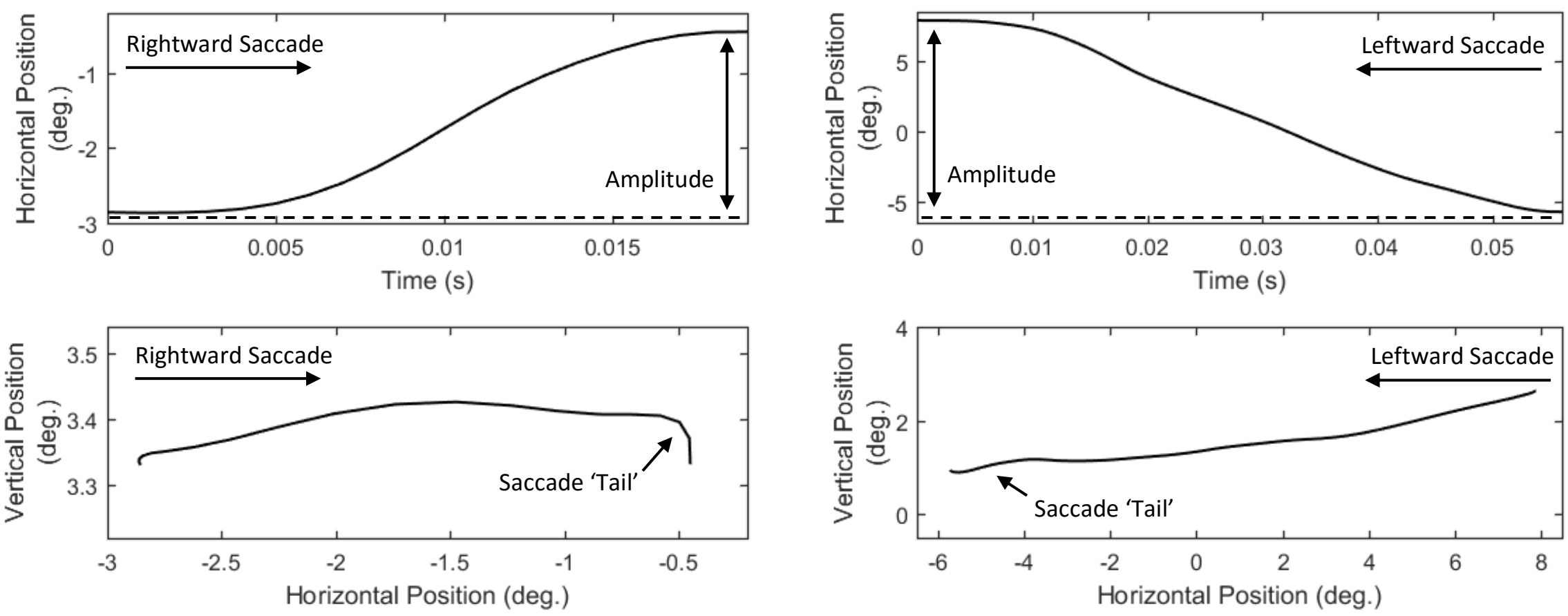

**Fig. 4** Characteristics of saccades in time domain (top) and in XY-plane/2-D trajectory (bottom).

**List 2.3.2** Saccade amplitude and curvature features

$\boldsymbol{S03}$: $\boldsymbol{SacAmp^{DistrStat-HVR}}$
$DistrStat(\cdot)$ on amplitudes of saccades: $SacAmp_i = |SacPos_i(end) - SacPos_i(start)|, i = 1, \dots, Sac^{Num}$

$\boldsymbol{S04}$: $\boldsymbol{SacTravDist^{DistrStat-R}}$
$DistrStat(\cdot)$ on travelled distances of saccades: $SacTravDist_i = \sum_{j=1}^{N-1}|SacPos_i(j+1) - SacPos_i(j)|$, $i = 1, \dots, Sac^{Num}$

$\boldsymbol{S05}$: $\boldsymbol{SacEfficiency^{DistrStat-R}}$
$DistrStat(\cdot)$ on efficiency metric of saccades: $SacEfficiency_i = \frac{SacAmp_i}{SacTravDist_i}, i = 1, \dots, Sac^{Num}$

$\boldsymbol{S06}$: $\boldsymbol{SacTailEfficiency^{DistrStat-R}}$
$DistrStat(\cdot)$ on tail efficiency metric of saccades: $SacTailEfficiency_i = \frac{SacTailAmp_i}{SacTailTravDist_i}, i = 1, \dots, Sac^{Num}$, where ‘Tails’ are defined as the samples of the last 7ms of a saccade

$\boldsymbol{S07}$: $\boldsymbol{SacTailPrInconsist^{DistrStat-HV2D}}$
$DistrStat(\cdot)$ on percentage tail inconsistency metric of saccades: $SacTailPrInconsist_i$, $i = 1, \dots, Sac^{Num}$ is the percentage of saccade ‘Tail’ for which $angle(LD_i, OD_i) \geq 60°$, where $LD_i$ the vector connecting the current and the previous point of a saccade, and $OD_i$ the vector connecting the starting and ending point of a saccade (raw signal was used)

$\boldsymbol{S08}$: $\boldsymbol{SacInitDir^{DistrStat-HV2D}}$

$DistrStat(\cdot)$ on initial direction of saccades: $SacInitDir_i = angle(ID_i, OD_i), i = 1, \dots, Sac^{Num}$, where $ID_i$ the vector connecting the starting point of a saccade and a predefined point (20ms afterwards), and $OD_i$ the vector connecting the starting and ending points of a saccade (in x-y plane)

$\boldsymbol{S09}$: $\boldsymbol{SacInitAvgDev^{DistrStat-HV2D}}$
$DistrStat(\cdot)$ on initial average deviation of saccades: $SacInitAvgDev_i = \sum_{j=1}^{m} InitDev_i(j)\ , i = 1, \dots, Sac^{Num}$, where $m$ the samples in a window of 10ms after the start of a saccade. $InitDev_i(j)$ is calculated by subtracting the eye position (of sample *j*) on the dimension orthogonal to the saccade direction from the value on that dimension at the start of the saccade

$\boldsymbol{S10}, \boldsymbol{S11}$: $\boldsymbol{SacMaxRawDev^{DistrStat-HV2D}}, \boldsymbol{SacPoiMaxRawDev^{DistrStat-HV2D}}$
$DistrStat(\cdot)$ on maximum raw deviation of saccades and the respective point: $SacMaxCurv_i = \max_{j=1,\dots,N} |PerpDist(j)|\ , i = 1, \dots, Sac^{Num}$, where $PerpDist(j)$ is the perpendicular distance (deviation) of sample $j$ from the straight line connecting the starting and ending points of a saccade. The respective point is expressed as percentage of the total duration of a saccade

$\boldsymbol{S12}$: $\boldsymbol{SacAreaCurv^{DistrStat-HV2D}}$
$DistrStat(\cdot)$ on area curvature metric of saccades: $SacAreaCurv_i = \sum_{j=2}^{N} SD(j) \cdot PD(j)\ , i = 1, \dots, Sac^{Num}$, where $SD(j)$ the distance covered by sample *j* along the straight path between onset and endpoint since the previous sample (*j-1*), and $PD(j)$ is the perpendicular (signed) deviation of sample *j*

$\boldsymbol{S13}$: $\boldsymbol{SacQuadCurv^{DistrStat-HV2D}}$
$DistrStat(\cdot)$ on quadratic-fit curvature metric of saccades: $SacQuadCurv_i\ i = 1, \dots, Sac^{Num}$ is the quadratic coefficient calculated via quadratic fitting on saccade position points*.

$\boldsymbol{S14}, \boldsymbol{S15}$: $\boldsymbol{SacCubCurvM1^{DistrStat-HV2D}}, \boldsymbol{SacPoiCubCurvM1^{DistrStat-HV2D}}$
$DistrStat(\cdot)$ on cubic-fit-extreme-1 of saccades and the respective point: $SacCubCurvM1_i\ i = 1, \dots, Sac^{Num}$ is the maximum of the cubic function fitted on the position points of a saccade*. The respective point is expressed as percentage of the total duration of a saccade

$\boldsymbol{S16}, \boldsymbol{S17}$: $\boldsymbol{SacCubCurvM2^{DistrStat-HV2D}}, \boldsymbol{SacPoiCubCurvM2^{DistrStat-HV2D}}$
$DistrStat(\cdot)$ on cubic-fit-extreme-2 of saccades and the respective point: $SacCubCurvM2_i\ i = 1, \dots, Sac^{Num}$ is the minimum of the cubic function fitted on the position points of a saccade*. The respective point is expressed as percentage of the total duration of a saccade

$\boldsymbol{S18}, \boldsymbol{S19}$: $\boldsymbol{SacCubCurvMax^{DistrStat-HV2D}}, \boldsymbol{SacPoiCubCurvMax^{DistrStat-HV2D}}$
$DistrStat(\cdot)$ on cubic-fit-curvature-maximum of saccades and the respective point: $SacCubCurvMax_i = max(SacCubCurvM1_i, SacCubCurvM2_i), SacPoiCubCurvMax_i = max(SacPoiCubCurvM1_i, SacPoiCubCurvM2_i), i = 1, \dots, Sac^{Num}$

### *2.3.3. Features of saccade velocity and acceleration*

Saccades are considered to be of ballistic nature and it is assumed that their velocity cannot be modulated intentionally (Becker & Fuchs, 1969). Thus the dynamic features of saccades provide an valuable source for exploring the background neurophysiological activity. In previous studies, they have been investigated as indicators of (de-)activation (Galley, 1989) and arousal (Di Stasi, Catena, Cañas, Macknik, & Martinez-Conde, 2013). In Fig. 5, we show examples of saccade velocity profiles and their characteristics. The feature of peak velocity (*S21*, List 2.3.3.1) is one of the basic dynamic features that can be easily extracted from the velocity profiles. In order to further model the characteristics of velocity profiles we extract the respective profile-modeling features (*S22*-*S26*, List 2.3.3.1).

* For this feature every saccade is translated so that the axis through its starting and ending positions coincides with the abscissa. The horizontal axis is rescaled so that each saccade starts at -1 and ends at +1

The acceleration of saccades is directly related to the force that moves the eyeball and thus it can provide important clues for the dynamic properties of saccades. The existence of asymmetries in the shapes of saccadic acceleration and deceleration phases has been reported by (Fricker, 1971) and it has been shown that their characteristics can be modulated by motor learning (Collins, Semroud, Orriols, & Doré-Mazars, 2008). Furthermore, abnormal characteristics of the saccade acceleration-deceleration phases have been reported in studies of autism spectrum disorders (Schmitt, Cook, Sweeney, & Mosconi, 2014). In Fig. 6, we show examples of saccade acceleration profiles (same saccades as in Fig. 5) demonstrating the peaks, durations and shapes of the acceleration-deceleration phases. In List 2.3.3.2 we describe the features extracted to model the acceleration characteristics of saccades (*S27*-*S33*).

**List 2.3.3.1** Saccade velocity features

$\boldsymbol{S20}$: $\boldsymbol{SacNumVelLocMin^{DistrStat-R}}$
$DistrStat(\cdot)$ on number of local minima in velocity profile of saccades: $SacNumVelLocMin_i$ is the number of sign changes from negative to positive in vector $SignVel(j) = sign(SacVel_i(j) - SacVel_i(j-1)), j = 2, \dots, N, i = 1, \dots, Sac^{Num}$

$\boldsymbol{S21}$: $\boldsymbol{SacPkVel^{DistrStat-HVR}}$
$DistrStat(\cdot)$ on peak velocities of saccades: $SacPkVel_i = \max_{j=1,\dots,N} |SacVel_i(j)|, i = 1, \dots, Sac^{Num}$

$\boldsymbol{S22}$: $\boldsymbol{SacVelProfMn^{DistrStat-HVR}}$
$DistrStat(\cdot)$ on velocity profile-sample mean of saccades: $SacVelProfMn_i = \sum_{j=1}^{N} |SacVel_i(j)|/N, i = 1, \dots, Sac^{Num}$

$\boldsymbol{S23}$: $\boldsymbol{SacVelProfMd^{DistrStat-HVR}}$
$DistrStat(\cdot)$ on velocity profile-sample median of saccades: $SacVelProfMd_i = median(|SacVel_i|), i = 1, \dots, Sac^{Num}$

$\boldsymbol{S24}$: $\boldsymbol{SacVelProfSd^{DistrStat-HVR}}$
$DistrStat(\cdot)$ on velocity profile-sample standard deviation of saccades: $SacVelProfSd_i = \sqrt{\sum_{j=1}^{N} (|SacVel_i(j)| - SacVelProfMn_i)^2/N}, i = 1, \dots, Sac^{Num}$

$\boldsymbol{S25}$: $\boldsymbol{SacVelProfSk^{DistrStat-HVR}}$
$DistrStat(\cdot)$ on velocity profile-sample skewness of saccades: $SacVelProfSk_i = \frac{\sum_{j=1}^{N} (|SacVel_i(j)| - SacVelProfMn_i)^3/N}{\left(\sqrt{\sum_{j=1}^{N} (|SacVel_i(j)| - SacVelProfMn_i)^2/N}\right)^3}, i = 1, \dots, Sac^{Num}$

$\boldsymbol{S26}$: $\boldsymbol{SacVelProfKu^{DistrStat-HVR}}$
$DistrStat(\cdot)$ on velocity profile-sample kurtosis of saccades: $SacVelProfKu_i = \frac{\sum_{j=1}^{N} (|SacVel_i(j)| - SacVelProfMn_i)^4/N}{\left(\sum_{j=1}^{N} (|SacVel_i(j)| - SacVelProfMn_i)^2/N\right)^2}, i = 1, \dots, Sac^{Num}$

**List 2.3.3.2** Saccade acceleration features

$\boldsymbol{S27}$: $\boldsymbol{SacPkAcc^{DistrStat-HVR}}$
$DistrStat(\cdot)$on peak accelerations of saccades: $SacPkAcc_i = \max_{j=1,\dots,idx-1} |SacAcc_i(j)|, i = 1, \dots, Sac^{Num}$, where $idx$ is the sample where $SacPkVel_i$ occurs

$\boldsymbol{S28}$: $\boldsymbol{SacPkDec^{DistrStat-HVR}}$
$DistrStat(\cdot)$ on peak decelerations of saccades: $SacPkDec_i = \max_{j=idx+1,\dots,N} |SacAcc_i(j)|, i = 1, \dots, Sac^{Num}$, where $idx$ is the sample where $SacPkVel_i$ occurs

$\boldsymbol{S29}$: $\boldsymbol{SacAccProfMn^{DistrStat-HVR}}$

$DistrStat(\cdot)$ on acceleration profile-sample mean of saccades: $SacAccProfMn_i = \sum_{j=1}^{N} |SacAcc_i(j)|/N, i = 1, \ldots, Sac^{Num}$

***S*30**: $\boldsymbol{SacAccProfMd^{DistrStat-HVR}}$
$DistrStat(\cdot)$ on acceleration profile-sample median of saccades: $SacAccProfMd_i = median(|SacAcc_i(j)|), i = 1, \ldots, Sac^{Num}$

***S*31**: $\boldsymbol{SacAccProfSd^{DistrStat-HVR}}$
$DistrStat(\cdot)$ on acceleration profile-sample standard deviation of saccades: $SacAccProfSd_i = \sqrt{\sum_{j=1}^{N} (|SacAcc_i(j)| - SacAccProfMn_i)^2/N}, i = 1, \ldots, Sac^{Num}$

***S*32**: $\boldsymbol{SacAccProfSk^{DistrStat-HVR}}$
$DistrStat(\cdot)$ on acceleration profile-sample skewness of saccades: $SacAccProfSk_i = \frac{\sum_{j=1}^{N} (|SacAcc_i(j)| - SacAccProfMn_i)^3/N}{\sqrt{\sum_{j=1}^{N} (|SacAcc_i(j)| - SacAccProfMn_i)^2/N}^3}, i = 1, \ldots, Sac^{Num}$

***S*33**: $\boldsymbol{SacAccProfKu^{DistrStat-HVR}}$
$DistrStat(\cdot)$ on acceleration profile-sample kurtosis of saccades: $SacAccProfKu_i = \frac{\sum_{j=1}^{N} (|SacAcc_i(j)| - SacAccProfMn_i)^4/N}{\left(\sum_{j=1}^{N} (|SacAcc_i(j)| - SacAccProfMn_i)^2/N\right)^2}, i = 1, \ldots, Sac^{Num}$

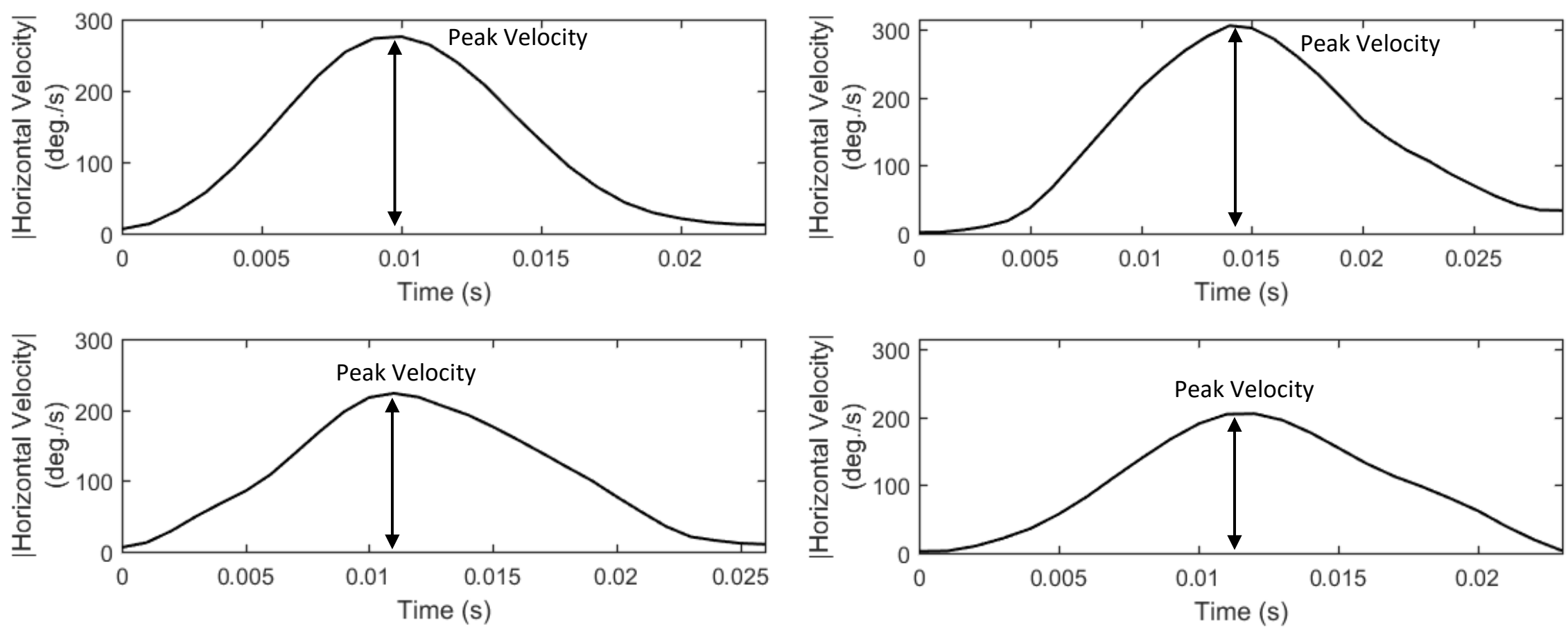

**Fig. 5** Typical saccade velocity profiles and their characteristics.

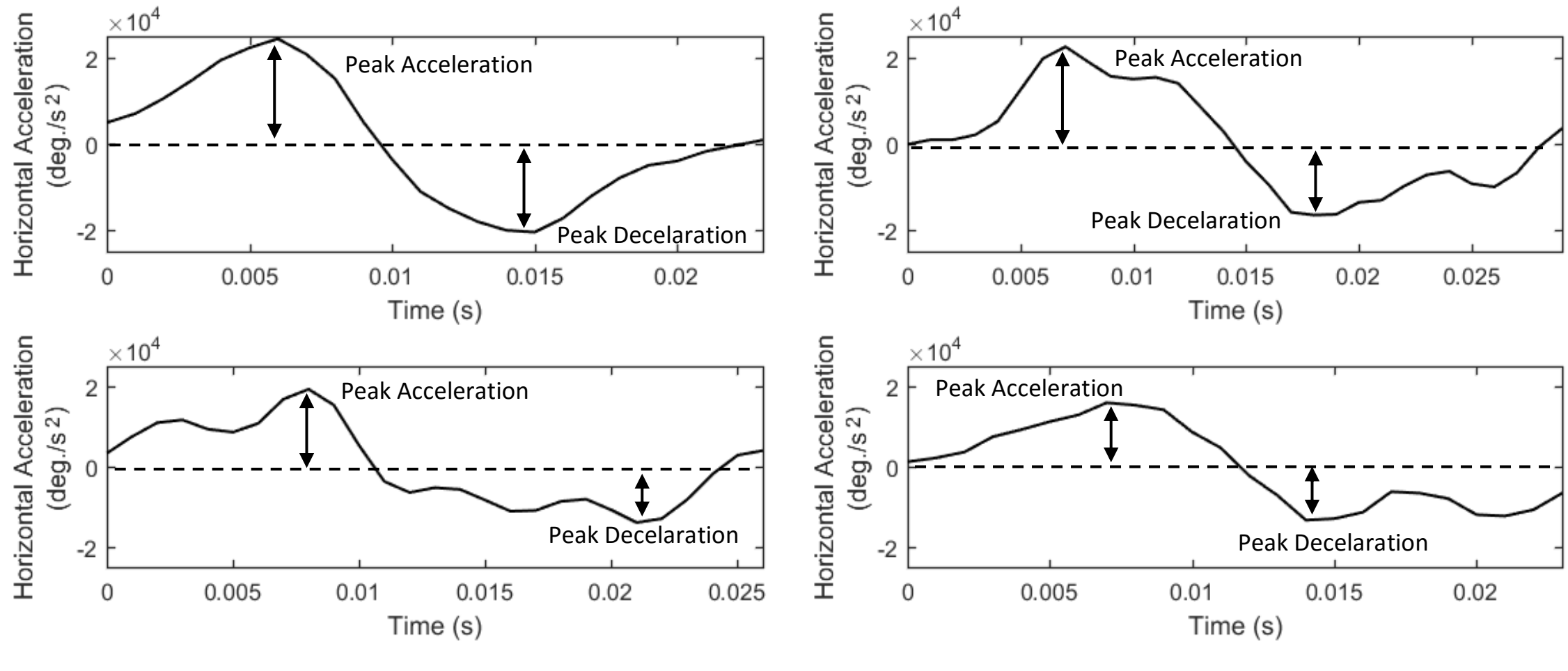

**Fig. 6** Saccade acceleration profiles demonstrating the differences in peak values, durations, and overall shapes of the acceleration-deceleration phases.

*2.3.4. Features of saccade-characteristic ratios*

The calculation of ratio features can provide useful information about possible inter-relationships of various oculomotor mechanisms, and also, such features can be used to provide robustness against exogenous effects, when these effects are not desired (e.g. effects from specific stimulus layout). Various ratio features have been investigated in the past in studies of the Parkinson's disease (peak velocity-mean velocity ratio or Q-ratio) (Garbutt, Harwood, Kumar, Han, & Leigh, 2003), as indicators of alertness (peak velocity-duration ratio or Saccadic-ratio) (Gupta & Routray, 2012), and in biometrics (peak acceleration-peak deceleration ratio) (Rigas, Komogortsev, & Shadmehr, 2016). In List 2.3.4 we present the characteristic ratio features (*S34*-*S40*) extracted from saccades.

**List 2.3.4** Saccade-characteristic ratio features

$\boldsymbol{S34}$: $\boldsymbol{SacAmpDur_{Ratio}^{DistrStat-HVR}}$
$DistrStat(\cdot)$ on amplitude-duration ratio of saccades: $SacAmpDur_{Ratio_i} = \frac{SacAmp_i}{SacDur_i}$ $i = 1, \dots, Sac^{Num}$

$\boldsymbol{S35}$: $\boldsymbol{SacPkVelAmp_{Ratio}^{DistrStat-HVR}}$
$DistrStat(\cdot)$ on peak velocity-amplitude ratio of saccades: $SacPkVelAmp_{Ratio_i} = \frac{SacPkVel_i}{SacAmp_i}$ $i = 1, \dots, sac^{Num}$

$\boldsymbol{S36}$: $\boldsymbol{SacPkVelDur_{Ratio}^{DistrStat-HVR}}$
$DistrStat(\cdot)$ on peak velocity-duration ratio of saccades: $SacPkVelDur_{Ratio_i} = \frac{SacPkVel_i}{SacDur_i}$ $i = 1, \dots, Sac^{Num}$

$\boldsymbol{S37}$: $\boldsymbol{SacPkVelMnVel_{Ratio}^{DistrStat-HVR}}$
$DistrStat(\cdot)$ on peak velocity-mean velocity ratio of saccades: $SacPkVelMnVel_{Ratio_i} = \frac{SacPkVel_i}{SacAmpDur_{Ratio_i}}$ $i = 1, \dots, Sac^{Num}$

$\boldsymbol{S38}$: $\boldsymbol{SacPkVelLocNoiseRatio^{DistrStat-R}}$
$DistrStat(\cdot)$ on peak velocity-local noise ratio of saccades: $SacPkVelLocNoise_{Ratio_i} = \frac{SacPkVel_i}{SacLocNoise_i}$ $i = 1, \dots, Sac^{Num}$, where $SacLocNoise_i$ is calculated adaptively using the velocity samples preceding a saccade

$\boldsymbol{S39}$: $\boldsymbol{SacAccDecDur_{Ratio}^{DistrStat}}$
$DistrStat(\cdot)$ on acceleration-deceleration duration ratio of saccades: $SacAccDecDur_{Ratio_i} = \frac{SacAcc_i^{EndTime} - SacAcc_i^{StartTime}}{SacDec_i^{EndTime} - SacAcc_i^{StartTime}}$ $i = 1, \dots, Sac^{Num}$, where $SacAcc_i^{StartTime}, SacAcc_i^{EndTime}, SacDec_i^{StartTime}, SacDec_i^{EndTime}$ are the starting and ending times of the acceleration and decelerations phases of a saccade

$\boldsymbol{S40}$: $\boldsymbol{SacPkAccPkDec_{Ratio}^{DistrStat-HVR}}$
$DistrStat(\cdot)$ on peak acceleration-peak deceleration ratio of saccades: $SacPkAccPkDec_{Ratio_i} = \frac{SacPkAcc_i}{SacPkDec_i}$ $i = 1, \dots, Sac^{Num}$

*2.3.5. Features of saccade main-sequence characteristics*

A more sophisticated way to describe the relationships between basic saccadic characteristics is to create a collective model (e.g. via curve fitting) using all the saccadic event instances in a recording. In the work of (Bahill et al., 1975) the general relationships of the feature pairs amplitude-duration and peak velocity-amplitude were investigated and the term 'main-sequence' (borrowed from Astronomy) was used to describe them. The characteristics of main-sequence have been investigated in relation to metal workload and arousal (Di Stasi, Antolí, & Cañas, 2011; Di Stasi et al., 2013), and also, they have been employed in eye movement biometrics (Rigas et al., 2016) in order to model saccadic vigor. The features extracted to describe the main-sequence relationships (*S41*-*S46*, List 2.3.5) were modeled by fitting linear curves directly on the amplitude-duration data, and on the logarithms of peak velocity-amplitude data (due to the existing non-linear relationship). In Fig. 7, we show examples of the performed curve fitting of the 'main sequence' relationships for the saccades extracted from a recording.

**List 2.3.5** Saccade main-sequence features

$\boldsymbol{S41}$: $\boldsymbol{SacAmpDurFitLn^{R}_{Intercept}}$
The intercept from the linear-regression-fit performed collectively on all saccades to model the overall amplitude-duration relationship $y = f(x)$, where $y = SacAmp_i, x = SacDur_i, i = 1, \dots, Sac^{Num}$

$\boldsymbol{S42}$: $\boldsymbol{SacAmpDurFitLn^{R}_{Slope}}$
The slope from the linear-regression-fit performed collectively on all saccades to model the overall amplitude-duration relationship $y = f(x)$, where $y = SacAmp_i, x = SacDur_i, i = 1, \dots, Sac^{Num}$

$\boldsymbol{S43}$: $\boldsymbol{SacAmpDurFitLn^{R}_{R^2}}$
The $R^2$ from the linear-regression-fit performed collectively on all saccades to model the overall amplitude-duration relationship $y = f(x)$, where $y = SacAmp_i, x = SacDur_i, i = 1, \dots, Sac^{Num}$

$\boldsymbol{S44}$: $\boldsymbol{SacPkVelAmpFitLn^{R}_{Intercept}}$
The intercept from the linear-regression-fit performed collectively on all saccades to model the overall logarithm peak velocity-logarithm amplitude relationship $y = f(x)$, where $y = \log(SacPkVel_i)\,, x = \log(SacAmp_i)\,, i = 1, \dots, Sac^{Num}$

$\boldsymbol{S45}$: $\boldsymbol{SacPkVelAmpFitLn^{R}_{Slope}}$
The slope from the linear-regression-fit performed collectively on all saccades to model the overall logarithm peak velocity-logarithm amplitude relationship $y = f(x)$, where $y = \log(SacPkVel_i)\,, x = \log(SacAmp_i)\,, i = 1, \dots, Sac^{Num}$

$\boldsymbol{S46}$: $\boldsymbol{SacPkVelAmpFitLn^{R}_{R^2}}$
The $R^2$ from the linear-regression-fit and performed collectively on all saccades to model the overall logarithm peak velocity-logarithm amplitude relationship $y = f(x)$, where $y = \log(SacPkVel_i)\,, x = \log(SacAmp_i)\,, i = 1, \dots, Sac^{Num}$

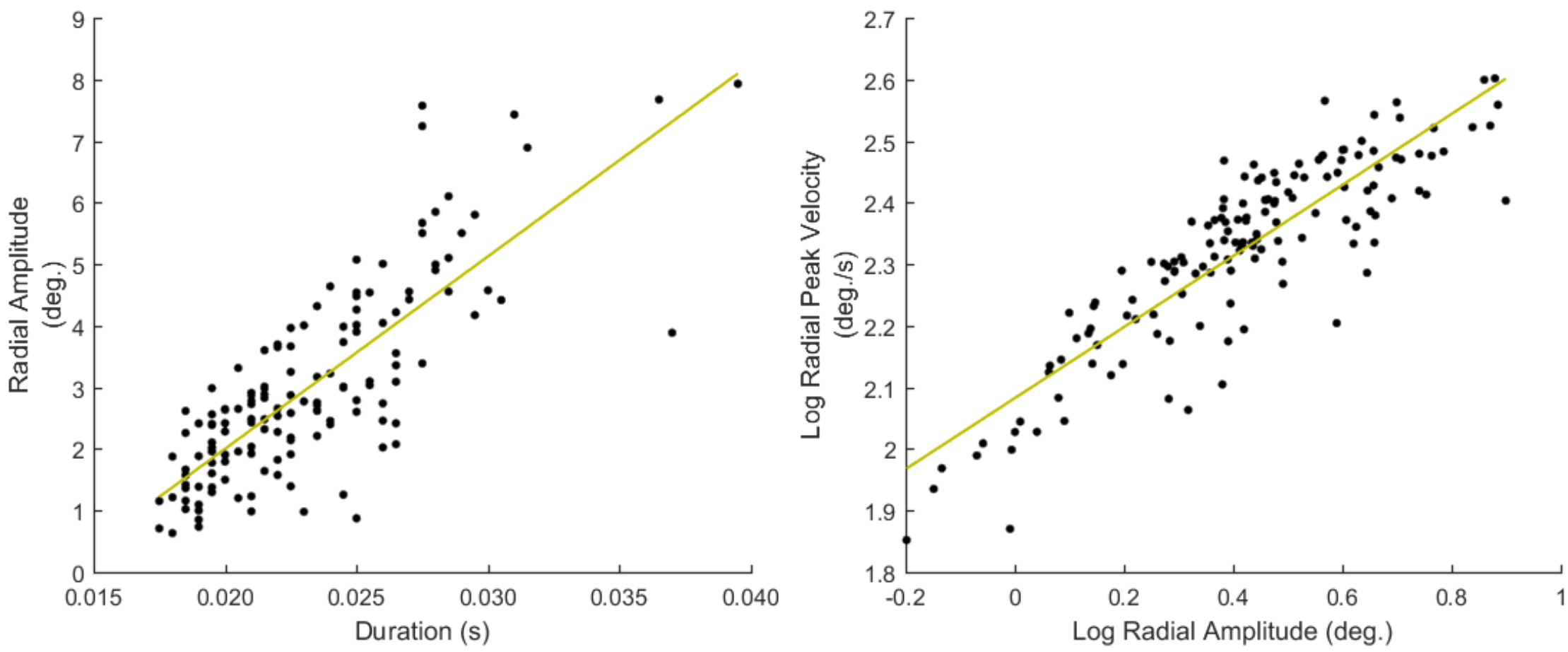

**Fig. 7** Saccade main-sequence relationships and the corresponding linear regression fit (left: amplitude-duration, right: peak velocity-amplitude).

*2.3.6. Features of saccade reading behavior*

The use of the reading paradigm allows for the extraction of a specialized group of saccadic features (*S47-S52*, List 2.3.6) that combine amplitude-direction cues and can represent saccadic events potentially connected to the reading behavior of subjects, e.g. forward read words, corrections and word regressions, line changes etc. Previous research on eye movements during reading has outlined the importance of such features and identified possible sources related to their variability (Rayner, 1998).

**List 2.3.6** Saccade reading behavior features

$\boldsymbol{S47}$: $\boldsymbol{SacSmRight_{Rate}}$
The number of small rightward saccades per second, i.e. saccades where $SacAmp_i^R \leq 8°$ and $SacPos_i^H(end) - SacPos_i^H(start) > 0, i = 1, \ldots, Sac^{Num}$

$\boldsymbol{S48}$: $\boldsymbol{SacSmLeft_{Rate}}$
The number of small leftward saccades per second, i.e. saccades where $SacAmp_i^R \leq 8°$ and $SacPos_i^H(end) - SacPos_i^H < 0, i = 1, \ldots, Sac^{Num}$

$\boldsymbol{S49}$: $\boldsymbol{SacLgRight_{Rate}}$
The number of large rightward saccades per second, i.e. saccades where $SacAmp_i^R > 8°$ and $SacPos_i^H(end) - SacPos_i^H(start) > 0, i = 1, \ldots, Sac^{Num}$

$\boldsymbol{S50}$: $\boldsymbol{SacLgLeft_{Rate}}$
The number of large leftward saccades per second, i.e. saccades where $SacAmp_i^R > 8°$ and $SacPos_i^H(end) - SacPos_i^H(start) > 0, i = 1, \ldots, Sac^{Num}$

$\boldsymbol{S51}$: $\boldsymbol{SacSmLeftSmRight_{Ratio}}$
The ratio of the number of small leftward saccades to the number of small and rightward saccades

$\boldsymbol{S52}$: $\boldsymbol{SacSmAllLgLeft_{Ratio}}$
The ratio of the number of all small saccades to the number of large and leftward saccades

## *2.4. Post-saccadic oscillation features*

A post-saccadic oscillation is a small oscillatory movement that occasionally can appear after a saccade. The term post-saccadic oscillation can cover movements appearing with various manifestations, e.g. as a small rapid movement, known as dynamic overshoot (Kapoula, Robinson, & Hain, 1986)), or as slower and smoother movement, known as glissadic overshoot (Weber & Daroff, 1972)). The exact sources and role of post-saccadic oscillations is not fully understood and their recording has been found to be pronounced when using specific eye-tracking technologies (Frens & van der Geest, 2002) or even influenced by the application of filtering. There are several studies that relate the appearance of glissadic phenomena with fatigue (Bahill & Stark, 1975b) and idiosyncratic characteristics (Kapoula et al., 1986). Irrespective of the exact origin and functionality of post-saccadic oscillations, the extraction of different features that model their characteristics can be useful in various fields of eye movement research.

*2.4.1. Features of post-saccadic oscillation temporal characteristics*

The basic features that are extracted to model the temporal characteristics of post-saccadic oscillations are the duration (*P01*) and two features modeling the frequency of appearance of post-saccadic oscillations, the interval between post-saccadic oscillations and the percent of saccades followed by a post-saccadic oscillation (*P02*-*P03*, List 2.4.1). To further quantify the frequency of appearance of the different manifestations of post-saccadic oscillations we extract features showing the percentages of slow, moderate, and fast post-saccadic oscillations (*P04*, *P05*, *P06*, List 2.4.1). The thresholds for the categorization of post-saccadic oscillations into slow, moderate and fast were selected after careful examination of their characteristics during the pre-processing stage (events classification, see Section 2.1).

**List 2.4.1** Post-saccadic oscillation temporal features

$\boldsymbol{P01}$: $\boldsymbol{PsoDur^{DistrStat}}$
$DistrStat(\cdot)$ on durations of post-saccadic oscillations: $PsoDur_i, i = 1, \ldots, Pso^{Num}$

$\boldsymbol{P02}$: $\boldsymbol{PsoInterv^{DistrStat}}$
$DistrStat(\cdot)$ on inter-post-saccadic oscillation intervals: $PsoInterv_i = Pso_i^{StartTime} - Pso_{i-1}^{EndTime}, i = 2, \ldots, Pso^{Num}$, where $Pso_i^{StartTime}, Pso_{i-1}^{EndTime}$ are the starting time of a post-saccadic oscillation and the ending time of the previous post-saccadic oscillation

$\boldsymbol{P03}$: $\boldsymbol{Pso_{Pr}}$
The percentage of saccades with a post-saccadic oscillation: $100\% \cdot (Pso^{Num}/Sac^{Num})$

$\boldsymbol{P04}$: $\boldsymbol{PsoSlow_{Pr}}$
The percentage of slow post-saccadic oscillations: $100\% \cdot (PsoSlow^{Num}/Pso^{Num})$, where $PsoSlow^{Num}$ is the number of slow post-saccadic oscillations, i.e. 20°/s < peak post-saccadic oscillation velocity < 45°/s

$\boldsymbol{P05}$: $\boldsymbol{PsoModerate_{Pr}}$
The percentage of moderate post-saccadic oscillations $100\% \cdot (PsoModerate^{Num}/Pso^{Num})$, where $PsoModerate^{Num}$ is the number of moderate post-saccadic oscillations, i.e. 45°/s < peak post-saccadic oscillation velocity < 55°/s

$\boldsymbol{P06}$: $\boldsymbol{PsoFast_{Pr}}$
The percentage of fast post-saccadic oscillations $100\% \cdot (PsoFast^{Num}/Pso^{Num})$, where $PsoFast^{Num}$ is the number of fast post-saccadic oscillations, i.e. peak post-saccadic oscillation velocity > 55°/s

*2.4.2. Features of post-saccadic oscillation shape*

Instead of extracting the 'amplitude' of post-saccadic oscillations (absolute difference between starting and ending positions), which is probably less informative due to their 'oscillatory' shape, we extract the feature that represents the maximum absolute deviation in these 'oscillatory' shapes (*P07*, List 2.4.2). We further model the position profiles of post-saccadic oscillations by extracting two features that represent the existing number of local minima (valleys) and maxima (peaks) (*P08*-*P09*, List 2.4.2).

In Fig. 8, we present examples of post-saccadic oscillation position profiles demonstrating the variability in their 'oscillatory' shapes and portraying the modeled features.

**List 2.4.2** Post-saccadic oscillation shape features

$\boldsymbol{P07}$: $\boldsymbol{PsoMaxAbsDev^{DistrStat-HVR}}$
$DistrStat(\cdot)$ on maximum absolute deviation of position profiles of post-saccadic oscillations: $PsoMaxAbsDev_i = \left|\max_j PsoPos_i(j) - \min_j PsoPos_i\,(j)\right|, i = 1, \dots, Pso^{Num}$

$\boldsymbol{P08}$: $\boldsymbol{PsoNumPosVlls^{DistrStat-HVR}}$
$DistrStat(\cdot)$ on number of valleys in position profile of post-saccadic oscillations: $PsoNumPosVlls_i$ is the number of sign changes from negative to positive in vector $SignPos(j) = sign(|PsoPos_i(j)| - |PsoPos_i(j-1)|), j = 2, \dots, N, i = 1, \dots, Pso^{Num}$

$\boldsymbol{P09}$: $\boldsymbol{PsoNumPosPks^{DistrStat-HVR}}$
$DistrStat(\cdot)$ on number of peaks in position profile of post-saccadic oscillations: $PsoNumPosPks_i$ is the number of sign changes from positive to negative in vector $SignPos(j) = sign(|PsoPos_i(j)| - |PsoPos_i(j-1)|), j = 2, \dots, N, i = 1, \dots, Pso^{Num}$

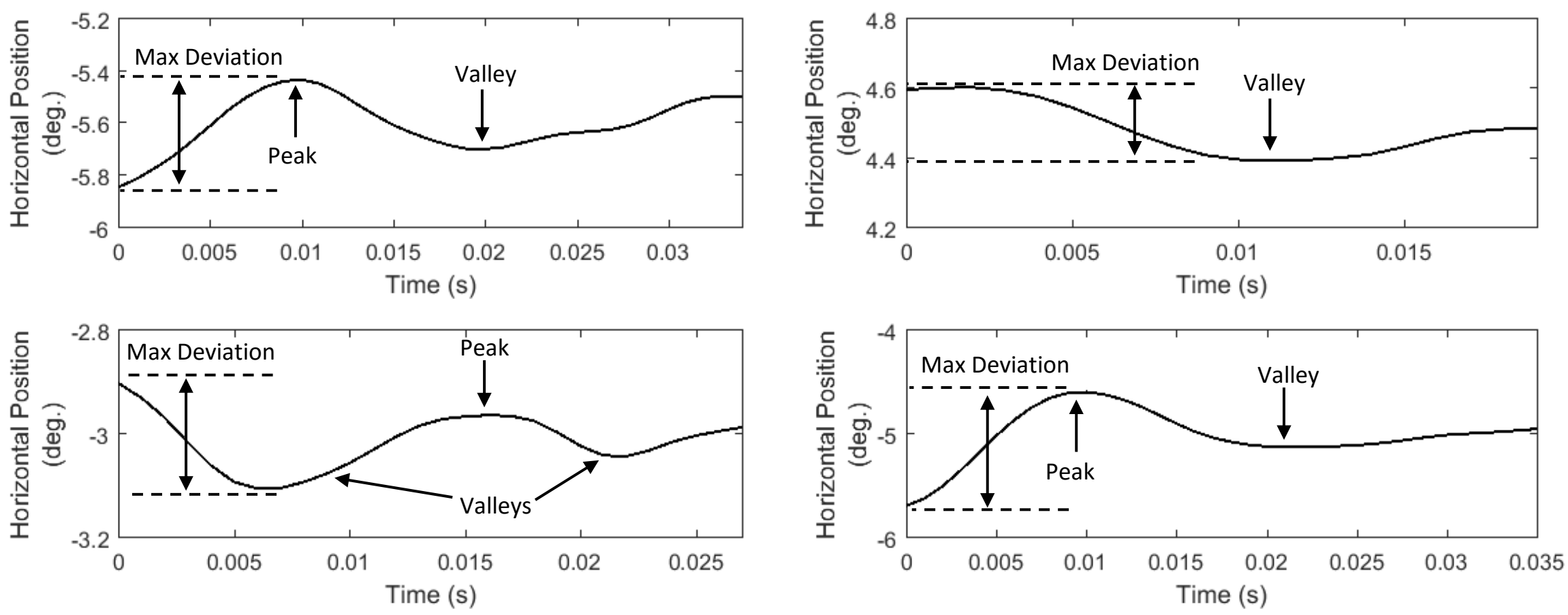

**Fig. 8** Post-saccadic oscillation position profiles with different 'oscillatory' shapes.

*2.4.3. Features of post-saccadic oscillation velocity and acceleration*

Due to the nature of post-saccadic oscillations their velocity profiles usually have multiple peaks. We extract the feature of peak velocity (*P09*, List 2.4.3.1) as the largest of them –most often is the first peak. In Fig. 9 we show examples of the velocity profiles of post-saccadic oscillations (corresponding to the same events as in Fig. 8). We demonstrate the differences in the peak velocities of post-saccadic oscillations that determine their characterization as fast, moderate or slow. In Fig. 10 we present the corresponding acceleration profiles. As performed for the events of fixations and saccades, we also extract a group of features that model the velocity and acceleration profiles of post-saccadic oscillations using descriptive statistics (*P11*-*P20*, List 2.4.3.1-2.4.3.2).

**List 2.4.3.1** Post-saccadic oscillation velocity features

$\boldsymbol{P10}$: $\boldsymbol{PsoPkVel^{DistrStat-HVR}}$
$DistrStat(\cdot)$ on peak velocities of post-saccadic oscillations: $PsoPkVel_i = \max_{j=1,\dots,N} |PsoVel_i(j)|, i = 1, \dots, Pso^{Num}$

$\boldsymbol{P11}$: $\boldsymbol{PsoVelProfMn^{DistrStat-HVR}}$
$DistrStat(\cdot)$ on velocity profile-sample mean of post-saccadic oscillations: $PsoVelProfMn_i = \sum_{j=1}^{N} |PsoVel_i(j)|/N, i = 1, \dots, Pso^{Num}$

$\boldsymbol{P12}$: $\boldsymbol{PsoVelProfMd^{DistrStat-HVR}}$
$DistrStat(\cdot)$ on velocity profile-sample median of post-saccadic oscillations: $PsoVelProfMd_i = median(|PsoVel_i|), i = 1, \dots, Pso^{Num}$

$\boldsymbol{P13}$: $\boldsymbol{PsoVelProfSd^{DistrStat-HVR}}$
$DistrStat(\cdot)$ on velocity profile-sample standard deviation of post-saccadic oscillations: $PsoVelProfSd_i = \sqrt{\sum_{j=1}^{N} (|PsoVel_i(j)| - PsoVelProfMn_i)^2/N}, i = 1, \dots, Pso^{Num}$

$\boldsymbol{P14}$: $\boldsymbol{PsoVelProfSk^{DistrStat-HVR}}$
$DistrStat(\cdot)$ on velocity profile-sample skewness of post-saccadic oscillations: $PsoVelProfSk_i = \frac{\sum_{j=1}^{N} (|PsoVel_i(j)| - PsoVelProfMn_i)^3/N}{\left(\sqrt{\sum_{j=1}^{N} (|PsoVel_i(j)| - PsoVelProfMn_i)^2/N}\right)^3}, i = 1, \dots, Pso^{Num}$

$\boldsymbol{P15}$: $\boldsymbol{PsoVelProfKu^{DistrStat-HVR}}$
$DistrStat(\cdot)$ on velocity profile-sample kurtosis of post-saccadic oscillations: $PsoVelProfKu_i = \frac{\sum_{j=1}^{N} (|PsoVel_i(j)| - PsoVelProfMn_i)^4/N}{\left(\sum_{j=1}^{N} (|PsoVel_i(j)| - PsoVelProfMn_i)^2/N\right)^2}, i = 1, \dots, Pso^{Num}$

**List 2.4.3.2** Post-saccadic oscillation acceleration features

$\boldsymbol{P16}$: $\boldsymbol{PsoAccProfMn^{DistrStat-HVR}}$
$DistrStat(\cdot)$ on acceleration profile-sample mean of post-saccadic oscillations: $PsoAccProfMn_i = \sum_{j=1}^{N} |PsoAcc_i(j)|/N, i = 1, \dots, Pso^{Num}$

$\boldsymbol{P17}$: $\boldsymbol{PsoAccProfMd^{DistrStat-HVR}}$
$DistrStat(\cdot)$ on acceleration profile-sample median of post-saccadic oscillations: $PsoAccProfMd_i = median(|PsoAcc_i|), i = 1, \dots, Pso^{Num}$

$\boldsymbol{P18}$: $\boldsymbol{PsoAccProfSd^{DistrStat-HVR}}$
$DistrStat(\cdot)$ on acceleration profile-sample standard deviation of post-saccadic oscillations: $PsoAccProfSd_i = \sqrt{\sum_{j=1}^{N} (|PsoAcc_i(j)| - PsoAccProfMn_i)^2/N}, i = 1, \dots, Pso^{Num}$

$\boldsymbol{P19}$: $\boldsymbol{PsoAccProfSk^{DistrStat-HVR}}$
$DistrStat(\cdot)$ on acceleration profile-sample skewness of post-saccadic oscillations: $PsoAccProfSk_i = \frac{\sum_{j=1}^{N} (|PsoAcc_i(j)| - PsoAccProfMn_i)^3/N}{\left(\sqrt{\sum_{j=1}^{N} (|PsoAcc_i(j)| - PsoAccProfMn_i)^2/N}\right)^3}, i = 1, \dots, Pso^{Num}$

$\boldsymbol{P20}$: $\boldsymbol{PsoAccProfKu^{DistrStat-HVR}}$
$DistrStat(\cdot)$ on acceleration profile-sample kurtosis of post-saccadic oscillations: $PsoAccProfKu_i = \frac{\sum_{j=1}^{N} (|PsoAcc_i(j)| - PsoAccProfMn_i)^4/N}{\left(\sum_{j=1}^{N} (|PsoAcc_i(j)| - PsoAccProfMn_i)^2/N\right)^2}, i = 1, \dots, Pso^{Num}$

### *2.4.4. Features of saccade-post-saccadic oscillation characteristic ratios*

Given the fact that every post-saccadic oscillation can be tied to a preceding ‘parent’ saccade, we extract three additional categories of features that can be used to model the possible interrelationships between the characteristics of saccades and their adjacent post-saccadic oscillations. These features (*P21-P24*,

List 2.4.4) are extracted by computing the ratios between some important characteristics of post-saccadic oscillations and saccades, in specific, the duration, the amplitude (saccade) or deviation (post-saccadic oscillation), and the peak velocity.

**List 2.4.4** Saccade-post-saccadic oscillation characteristic ratio features

$\boldsymbol{P21}$: $\boldsymbol{PsoSDurPDur_{Ratio}^{DistrStat}}$
$DistrStat(\cdot)$ on saccade-post-saccadic oscillation duration ratios: $SDurPDur_{Ratio_i} = \frac{SacDur_i}{PsoDur_i}, i = 1, \dots, Pso^{Num}$
$\boldsymbol{P22}$: $\boldsymbol{PsoSAmpPDur_{Ratio}^{DistrStat-HVR}}$
$DistrStat(\cdot)$ on saccade amplitude-post-saccadic oscillation duration ratios: $SAmpPDur_{Ratio_i} = \frac{SacAmp_i}{PsoDur_i}, i = 1, \dots, Pso^{Num}$
$\boldsymbol{P23}$: $\boldsymbol{PsoSAmpPMaxAbsDev_{Ratio}^{DistrStat-HVR}}$
$DistrStat(\cdot)$ on saccade amplitude-post-saccadic oscillation maximum absolute deviation ratios: $SAmpPMaxAbsDev_{Ratio_i} = \frac{SacAmp_i}{PsoMaxAbsDev_i}, i = 1, \dots, Pso^{Num}$
$\boldsymbol{P24}$: $\boldsymbol{PsoSPkVelPPkVel_{Ratio}^{DistrStat-HVR}}$
$DistrStat(\cdot)$ on saccade-post-saccadic oscillation peak velocity ratios: $SPkVelPPkVel_{Ratio_i} = \frac{SacPkVel_i}{PsoPkVel_i}, i = 1, \dots, Pso^{Num}$

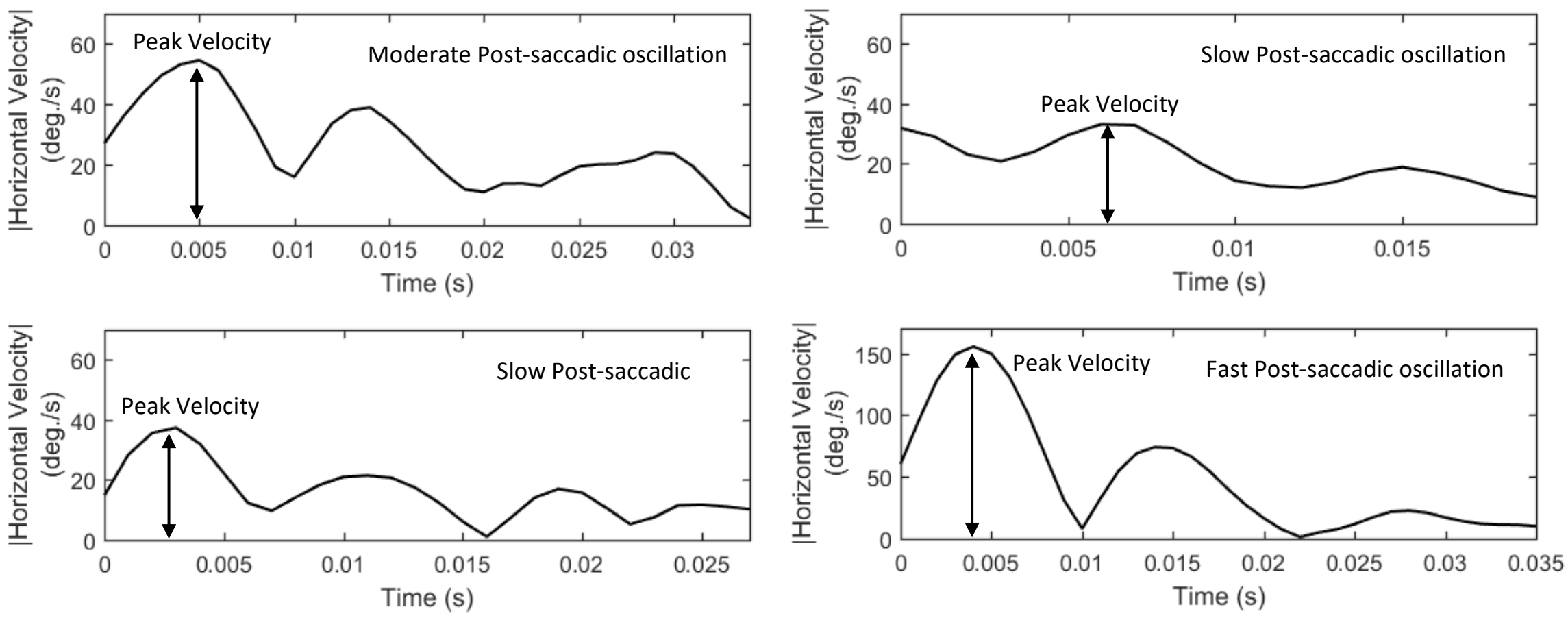

**Fig. 9** Post-saccadic oscillation velocity profiles and extraction of peak velocity.

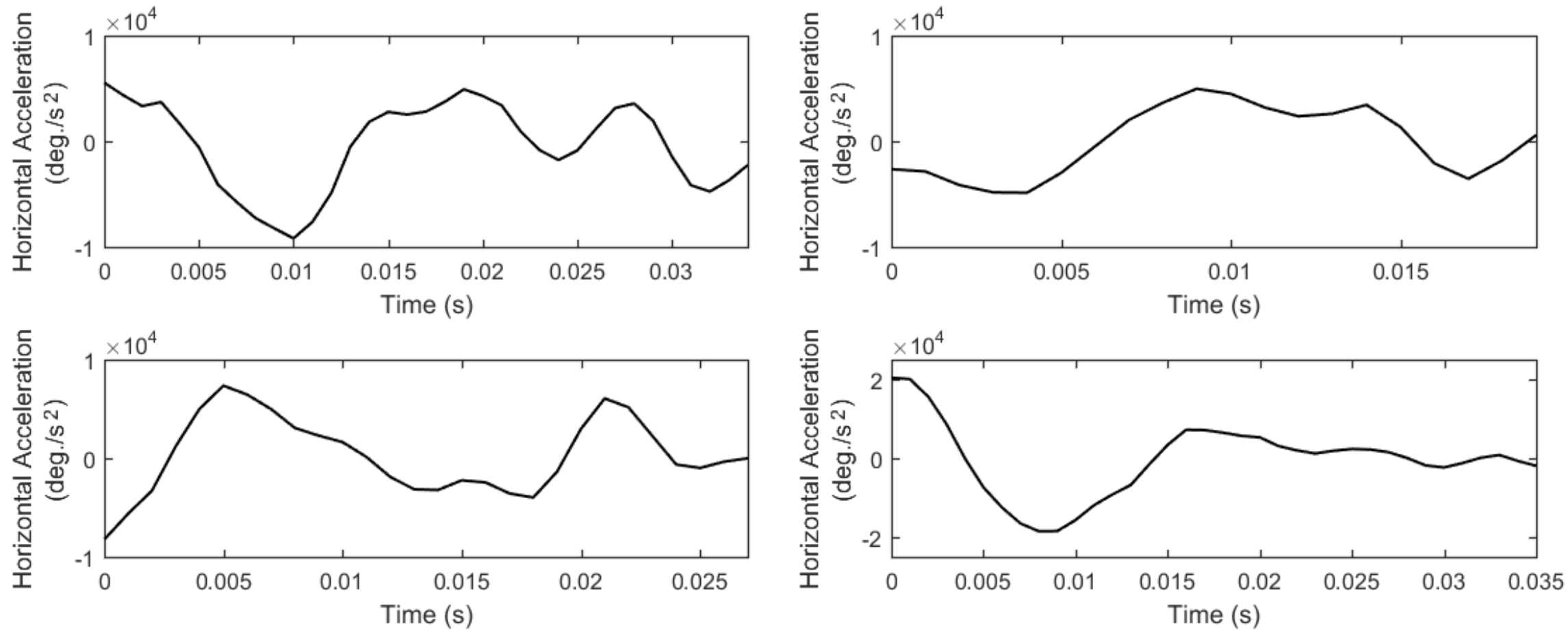

**Fig. 10** Post-saccadic oscillation acceleration profiles and their characteristics.

## 3. Experiments

### *3.1. Subjects*

The experiments were performed with the participation of 298 subjects (162 males/136 females) with ages from 18 to 46 years, (M = 22, SD = 4.3). All subjects had normal or corrected vision (147 normal /151 corrected with 61 glasses/86 contact lenses) and filled a questionnaire to verify that they did not have any recent severe head injury that could affect the oculomotor functionality. The study was approved by the institutional review board of Texas State University and the participants provided signed informed consent.

### *3.2. Apparatus and recording setup*

The eye tracking system used for the experiments was an EyeLink 1000 eye tracker* with a sampling rate of 1000 Hz. The eye tracker operated in monocular mode capturing the left eye. The typical vendor specifications of this system report accuracy of 0.5° and spatial resolution of 0.01° RMS. In our experiments we followed a strict protocol to ensure the high quality of recordings by restricting the allowed calibration accuracy error to maximum values lower than 1.5° and average values lower than 1°. We practically measured the average calibration accuracy over all recordings to be 0.48° (SD = 0.17°) and the average data validity to be 94.2% (SD = 5.7%). Validity is defined as the percentage of samples that were successfully captured by the eye-tracking device during a recording. Common sources of eye-tracking failure (invalidity) can be blinks, moisture, squinting etc. During the recordings, each subject was comfortably positioned at a distance of 550 mm from a computer screen with dimensions 474 × 297 mm and resolution 1680 × 1050 pixels, where the visual stimulus was presented. To mitigate any possible eye-tracking artifacts from small head movements, the subjects' heads were stabilized using a chin-rest with a forehead.

### *3.3. Experimental paradigm*

Every subject performed two recordings (two session test-retest paradigm) separated by an interval of 13 to 42 minutes (M = 19.5, SD = 4.2). In the time period between the two recordings the subjects performed various other eye movement tasks and had brief periods of rest to mitigate eye fatigue. The employed visual stimulus consisted from the six first stanzas of the poem of Lewis Carroll "The Hunting of the Snark". The participants were asked to freely read the text. The particular nonsense poem of Lewis Carroll was purposefully selected to encourage the active processing of text from the participants. The text excerpts from the poem were presented in white color on a black background using Times New Roman bold font of size 20pt, single-spaced (corresponding to 0.92° height of each line). The presented text covered a visual field of approximately 20° horizontally and 30° vertically.

* http://www.sr-research.com/EL_1000.html

## 4. Analysis

In this section, we present the performed analysis aiming to explore the typical values, the variability, and the test-retest reliability of the extracted features. For the calculation of summary statistics describing the central tendency and variability of features we employ the median and inter-quartile range over the feature values from the experimental population. We use these measures instead of the mean and standard deviation because these measures are expected to summarize the feature values more robustly both for normal and for non-normal features. We also employ two powerful measures to assess reliability, ICC and Kendall's W, used for the case of normally (the former) and non-normally (the latter) distributed features (see Section 4.1.2 for definitions). The normality is assessed by following the procedure described in Section 4.1.1. In our experimental paradigm the assessment of reliability is particularly important because it can serve two purposes: a) it reveals the stability of measurements taken on different occasions (sessions), and b) it can be used as a proxy of the relative variability of feature values between subjects (inter-subject) and sessions (intra-subject), and thus it can indicate the relative discriminatory power of each feature.

### *4.1. Analysis methods*

#### *4.1.1. Assessment of normality*

To assess the direct normality of features we employ the Pearson $\chi^2$ test and calculate the $p$ values at 5% significance level. Features that are not directly normal are subjected to a number of classic normalization transformations followed by reassessment of the test. The used transformations applied on each feature distribution $X$ are: the logarithm → $\log(X+1)$, the square root → $sqrt(X+0.5)$, the cube root → $sign(X)\cdot\sqrt[3]{|X|}$, the reciprocal → $1/X$, the arcsine → $2\cdot\sin^{-1}\sqrt{X}$ (for proportions), the logit → $\log adjX/(1 - adjX)$ (for proportions and features in range [0, 1]) with $adjX$ representing $X$ adjusted in range [0.025, 0.075] to avoid undefined cases. The logarithm, square root, cube root, and reciprocal were complementary evaluated using the reflection transformation → $max(X) + 1 - X$. Furthermore, in order to evaluate cases where the deviation from normality is due to outliers at the extremes, we perform Winsorization (Ruppert, 2004) with maximum-minimum limits at 5%-95% percentiles of distribution, and we reassess normality following the previous procedures.

*4.1.2. Assessment of reliability*

***Intraclass Correlation Coefficient (ICC)***: The ICC is a measure that can be calculated for normally distributed data in order to evaluate either the absolute agreement (accounts for systematic differences) or the consistency (does not account for systematic differences) of quantitative measurements, and thus, it can be used to assess the reliability among different raters (or, occasions). The work of (Shrout & Fleiss, 1979) describes the six basic forms of ICC and the procedures for their calculation. In the current work we use the ICC to assess absolute agreement, and for this reason, we employ the third ICC form (ICC(2, 1) in (Shrout & Fleiss, 1979) convention). The original approach for calculating the respective variance estimates was based on ANOVA tables. For cases involving two-way random effect models (like current case) there is also a more robust approach for this calculation based on variance component maximum likelihood (VCML) analysis (Searle, Casella, & McCulloch, 1992). We currently adopt this approach to calculate the ICC. The ICC takes values in range [0.00, 1.00] (1.00 indicates complete agreement). The work of (Cicchetti, 1994) suggested some rules of thumb for interpreting the ICC values, in specific, [0.75, 1.00] indicates 'excellent' agreement, [0.60, 0.75) indicates 'good' agreement, [0.40, 0.60) indicates 'fair' agreement, and [0.00, 0.40) indicates 'poor' agreement.

***Kendall's coefficient of concordance (Kendall's W)***: The Kendall's W (Kendall & Babington Smith, 1939) is a non-parametric measure that can be used to assess rater (or, occasion) agreement without the requirement for normally distributed data. The Kendall's W is calculated as a normalization of the Friedman test statistic (Friedman, 1937) in range [0.00, 1.00], with 1.00 indicating complete agreement. The process of estimating Kendall's W does not make any prior assumption for the nature of the data distribution but instead performs statistical calculations based on data rankings. Since Kendall's W is calculated for non-normally distributed data and since there are no similar rules (ranges) for the interpretation of values as for the ICC ('excellent', 'good' etc.), it is not advised to attempt a direct comparison of values between ICC and Kendall's W.

*4.2. Results and discussion*

The tables of results presented below are structured in two-levels: the top part presents single values for the non-distributional categories of features. The bottom part presents multiple values for the six subtypes of distributional categories of features (in corresponding columns). As already described, these values are extracted for each distributional feature category by calculating descriptive statistics (columns ***MN***, ***MD***, ***SD***, ***IQ***, ***SK***, ***KU***) on the multiple feature instances within a recording. The tables present values only for the independent components of eye movement (horizontal **H** and vertical **V**), except for the features extracted only for the radial component or from the trajectory in 2-D space. In Tables 4.2.1, 4.2.3 and 4.2.5 (fixations, saccades, and post-saccadic oscillations respectively) we present the values of central tendency (median, denoted **MD**) and overall variability (inter-quartile range, denoted **IQ**) for the features across all subjects' population. In Tables 4.2.2, 4.2.4, and 4.2.4 we present the respective measures for the assessment of normality and reliability of features. For each feature, one column indicates the maximum p value (**p)** calculated following the procedures described in Section 4.1.1, and the adjacent column presents the value of either the ICC when p value denotes a normally distributed feature ($p \geq 0.05$), or Kendall's W when p value denotes a non-normally distributed feature ($p < 0.05$). To further facilitate overview of results, the cells that correspond to non-normal features have been highlighted using light grey shading. Although the two reliability measures are presented interchangeably in the same column (**ICC/W**) to enable the compact overview of results, we should emphasize once more that it is not advised to directly compare the values of ICC and W.

In Table 4.2.1 we can overview the typical values of fixations features calculated over the used large experimental database. We can see that the median fixation duration was calculated to be about 200 ms (*F02*) and corresponds on an average rate of about 3-4 fixations per second (*F01*). This duration is within the expected range for fixations during reading, and similar values have been reported in previous research studies (Nyström & Holmqvist, 2010; Rayner, 1998). Since the fixation centroid (*F03*) is a direct measure of position, the extracted values for this feature are heavily affected by the positioning and centering of the stimulus.  However, when a common stimulus is used for all subjects (as in our experiments) the median and inter-quartile range can provide clues about the existence of systematic error and its variability, either system-related or subject-related (unique error signature) (Hornof & Halverson, 2002). As revealed from the values of *F05-F06*, the drift during fixation affects both components of eye movement similarly. Furthermore, the drift speeds of the two components (*F06*) are in very close proximity to previously reported values of 0.5°/s (Poletti et al., 2010). The values of the drift linear-fit slope feature (*F07*) reveal a positive tendency for the horizontal component and negative tendency for the vertical. Another important observation is that the values of the quadratic-fit $R^2$ feature (*F09*) are larger than these of the linear-fit $R^2$ feature (*F08*), which seems to further support the occasional appearance of non-linearity (curvature) in fixation drifts (see Fig. 1), a phenomenon previously reported in (Cherici, Kuang, Poletti, & Rucci, 2012). Finally, the calculated values for velocity and acceleration (*F14-F25*) demonstrate the exact levels of eye mobility during fixations, which expectedly are lower compared to the corresponding levels for saccades and post-saccadic oscillations (see corresponding tables).

The examination of Table 4.2.2 allows for a comparative assessment of the normality and reliability of fixation features. In overall, 50.7% of fixation features (feature subtypes) are found to be normally distributed and the rest are distributed non-normally. An examination of the shaded parts of the table (non-normal features) reveals that there is a general tendency for non-normality from the acceleration feature categories, and from the kurtosis (*KU*) descriptive statistic irrespective of feature category. For the case of fixations, the calculated ICC values for assessing reliability are in range of 0.06 to 0.92. Following the categorization suggested by (Cicchetti, 1994) we can see that 32.5% of them are in the region of 'excellent' reliability, 20.5% in the region of 'good' reliability, 23.9% in the region of 'fair' reliability, and 23.1% in the region of 'poor' reliability. When considering the feature categories in overall, the top performing features in terms of reliability are *F14-F16* (modeling of fixation velocity profile with mean, median, and standard deviation), *F09* ($R^2$ when modeling fixation drift with quadratic-fit), and *F02* (fixation duration). For the case of non-normal features, the calculated W values are in range of 0.52 to 0.98. This difference in ranges of ICC and W portrays the (already noted) risk to attempt a direct comparison of values from these two measures, since the values of W seem to be by-design compressed in the upper half of complete range [0.00, 1.00]. From the respective values for W we can see that generally the top performing feature categories based on Kendall's W measure are *F21-F23* (modeling of fixation acceleration profile with mean, median, and standard deviation), and *F05* (travelled distance during fixation drift). It is interesting to observe that although the eye mobility is relatively limited during fixations, the dynamic features (based on velocity and acceleration) seem to provide the best test-retest measurement agreement both for the case of normal and for non-normal features.

**Table 4.2.1** Statistics of central tendency and variability for fixation features over the experimental population.

**Non-Distributional Features**

| Feature Type | | MD | IQ | Feature Type | | MD | IQ | Feature Type | | MD | IQ |
|---|---|---|---|---|---|---|---|---|---|---|---|
| *F01 ($s^{-1}$)* | | 3.81 | 0.56 | *F11 (%)* | *H* | 7.58 | 3.52 | *F13 (%)* | *H* | 75.78 | 8.47 |
| | | | | | *V* | 8.37 | 3.49 | | *V* | 73.20 | 7.54 |
| *F10 (%)* | *H* | 9.25 | 4.57 | *F12 (%)* | *H* | 7.30 | 3.27 | | | | |
| | *V* | 9.47 | 4.18 | | *V* | 8.55 | 3.35 | | | | |

**Distributional Feature Subtypes**
**Descriptive Statistic: $DistrStat(\cdot)$**

| Feature Type | | *MN* | | *MD* | | *SD* | | *IQ* | | *SK** | | *KU** | |
|---|---|---|---|---|---|---|---|---|---|---|---|---|---|
| | | MD | IQ | MD | IQ | MD | IQ | MD | IQ | MD | IQ | MD | IQ |
| *F02 (ms)* | | 201.54 | 35.35 | 185.00 | 31.25 | 90.57 | 28.38 | 91.00 | 31.75 | 1.45 | 0.94 | 6.56 | 5.09 |
| *F03 (°)* | *H* | 0.09 | 0.82 | -0.07 | 1.05 | 4.72 | 0.34 | 7.97 | 0.75 | -0.13 | 0.22 | 2.42 | 0.58 |
| | *V* | -3.37 | 1.65 | -2.84 | 2.36 | 8.44 | 0.82 | 14.56 | 1.86 | -0.01 | 0.19 | 1.80 | 0.15 |
| *F04 (°)* | *H* | 0.12 | 0.04 | 0.09 | 0.03 | 0.11 | 0.05 | 0.12 | 0.04 | 1.97 | 1.20 | 8.45 | 7.68 |
| | *V* | 0.10 | 0.04 | 0.08 | 0.02 | 0.10 | 0.05 | 0.10 | 0.04 | 2.19 | 1.30 | 9.78 | 9.04 |
| *F05 (°)* | *H* | 0.94 | 0.49 | 0.83 | 0.44 | 0.49 | 0.30 | 0.45 | 0.28 | 1.83 | 1.03 | 8.00 | 6.26 |
| | *V* | 0.98 | 0.50 | 0.87 | 0.46 | 0.50 | 0.32 | 0.51 | 0.31 | 1.68 | 0.97 | 7.47 | 6.32 |
| *F06 (°/s)* | *H* | 0.65 | 0.23 | 0.50 | 0.18 | 0.58 | 0.26 | 0.65 | 0.24 | 1.81 | 0.94 | 7.50 | 5.64 |
| | *V* | 0.55 | 0.22 | 0.40 | 0.14 | 0.59 | 0.37 | 0.53 | 0.21 | 2.62 | 1.53 | 12.96 | 12.96 |
| *F07 (°/s)* | *H* | 0.12 | 0.18 | 0.09 | 0.21 | 0.82 | 0.32 | 0.87 | 0.34 | 0.26 | 0.98 | 5.79 | 3.60 |
| | *V* | -0.03 | 0.23 | -0.05 | 0.19 | 0.81 | 0.43 | 0.75 | 0.28 | 0.21 | 1.72 | 8.97 | 7.88 |
| *F08* | *H* | 0.44 | 0.11 | 0.44 | 0.17 | 0.30 | 0.02 | 0.54 | 0.08 | 0.04 | 0.41 | 1.73 | 0.20 |
| | *V* | 0.43 | 0.09 | 0.43 | 0.15 | 0.30 | 0.02 | 0.55 | 0.08 | 0.08 | 0.38 | 1.71 | 0.18 |
| *F09* | *H* | 0.59 | 0.11 | 0.65 | 0.16 | 0.26 | 0.02 | 0.40 | 0.06 | -0.52 | 0.49 | 2.19 | 0.55 |
| | *V* | 0.57 | 0.11 | 0.61 | 0.15 | 0.26 | 0.02 | 0.42 | 0.07 | -0.40 | 0.46 | 2.08 | 0.44 |
| *F14 (°/s)* | *H* | 3.44 | 1.47 | 3.25 | 1.40 | 0.77 | 0.44 | 0.85 | 0.42 | 1.35 | 1.13 | 5.86 | 5.17 |
| | *V* | 3.54 | 1.57 | 3.34 | 1.38 | 0.87 | 0.57 | 0.92 | 0.56 | 1.32 | 1.53 | 6.16 | 7.84 |
| *F15 (°/s)* | *H* | 2.74 | 1.22 | 2.60 | 1.18 | 0.66 | 0.40 | 0.75 | 0.37 | 1.33 | 1.17 | 6.10 | 5.23 |
| | *V* | 2.90 | 1.29 | 2.74 | 1.19 | 0.78 | 0.50 | 0.83 | 0.49 | 1.29 | 1.41 | 5.86 | 6.42 |
| *F16 (°/s)* | *H* | 2.91 | 1.15 | 2.67 | 1.07 | 0.98 | 0.46 | 0.84 | 0.39 | 2.44 | 1.62 | 11.26 | 10.34 |
| | *V* | 2.78 | 1.22 | 2.62 | 1.04 | 0.75 | 0.50 | 0.76 | 0.41 | 1.71 | 1.96 | 8.57 | 12.34 |
| *F17* | *H* | 1.26 | 0.24 | 1.11 | 0.15 | 0.66 | 0.29 | 0.57 | 0.24 | 2.09 | 1.04 | 9.41 | 6.57 |
| | *V* | 1.06 | 0.09 | 1.02 | 0.07 | 0.38 | 0.11 | 0.43 | 0.09 | 1.19 | 1.18 | 6.23 | 6.16 |
| *F18* | *H* | 4.99 | 1.26 | 3.94 | 0.47 | 3.61 | 2.32 | 2.13 | 1.06 | 3.51 | 1.60 | 18.75 | 14.90 |
| | *V* | 4.05 | 0.37 | 3.64 | 0.22 | 1.68 | 0.80 | 1.49 | 0.34 | 2.70 | 2.09 | 14.11 | 17.20 |
| *F19 (%)* | | 3.54 | 2.43 | 2.08 | 2.76 | 4.12 | 2.17 | 3.94 | 1.67 | 2.46 | 1.52 | 10.85 | 10.22 |
| *F20 (%)* | | 2.46 | 2.11 | 1.57 | 2.10 | 2.52 | 1.25 | 2.56 | 1.48 | 1.79 | 1.49 | 7.31 | 7.31 |
| *F21 (°/$s^2$)* | *H* | 970.59 | 532.93 | 916.30 | 530.15 | 192.69 | 180.26 | 199.52 | 149.80 | 0.92 | 1.92 | 5.11 | 7.56 |
| | *V* | 1081.4 | 543.5 | 1033.8 | 534.7 | 245.85 | 216.43 | 257.63 | 179.94 | 0.93 | 2.15 | 4.84 | 8.79 |
| *F22 (°/$s^2$)* | *H* | 778.92 | 425.21 | 735.89 | 427.30 | 171.21 | 150.92 | 178.35 | 130.06 | 0.85 | 1.42 | 4.78 | 5.62 |
| | *V* | 868.74 | 433.95 | 828.11 | 423.55 | 214.25 | 173.24 | 229.76 | 153.74 | 0.89 | 1.53 | 4.58 | 6.16 |
| *F23 (°/$s^2$)* | *H* | 793.72 | 443.25 | 745.84 | 422.62 | 167.78 | 164.06 | 172.51 | 124.06 | 1.11 | 2.05 | 5.48 | 9.92 |
| | *V* | 883.77 | 450.08 | 839.21 | 439.42 | 218.22 | 201.96 | 221.26 | 155.43 | 0.96 | 2.50 | 4.87 | 12.49 |
| *F24* | *H* | 1.19 | 0.06 | 1.15 | 0.05 | 0.35 | 0.10 | 0.41 | 0.08 | 1.03 | 1.01 | 5.30 | 4.13 |
| | *V* | 1.17 | 0.04 | 1.14 | 0.04 | 0.32 | 0.05 | 0.39 | 0.05 | 0.75 | 0.60 | 4.37 | 2.51 |
| *F25* | *H* | 4.51 | 0.39 | 4.13 | 0.21 | 1.68 | 0.85 | 1.61 | 0.33 | 2.27 | 1.61 | 11.00 | 12.01 |
| | *V* | 4.38 | 0.19 | 4.07 | 0.17 | 1.41 | 0.29 | 1.50 | 0.21 | 1.74 | 1.09 | 7.84 | 7.46 |

* Skewness and kurtosis are unit-less measures, so, the feature units do not apply on them

**Table 4.2.2** Statistics of normality and reliability for fixation features over the experimental population.

**Non-Distributional Features**

| Feature Type | | p | ICC/W | Feature Type | | p | ICC/W | Feature Type | | p | ICC/W |
|---|---|---|---|---|---|---|---|---|---|---|---|
| *F01* ($s^{-1}$) | | 0.39 | 0.82 | *F11 (%)* | *H* | 0.47 | 0.31 | *F13 (%)* | *H* | 0.21 | 0.68 |
| | | | | | *V* | 0.39 | 0.40 | | *V* | 0.22 | 0.64 |
| *F10 (%)* | *H* | 0.53 | 0.53 | *F12 (%)* | *H* | 0.47 | 0.36 | | | | |
| | *V* | 0.62 | 0.44 | | *V* | 0.21 | 0.37 | | | | |

**Distributional Feature Subtypes**
**Descriptive Statistic: $DistrStat(\cdot)$**

| Feature Type | | MN | | MD | | SD | | IQ | | SK | | KU | |
|---|---|---|---|---|---|---|---|---|---|---|---|---|---|
| | | p | ICC/W | p | ICC/W | p | ICC/W | p | ICC/W | p | ICC/W | p | ICC/W |
| *F02 (ms)* | | 0.34 | 0.86 | 0.14 | 0.86 | 0.05 | 0.68 | 0.40 | 0.78 | 0.05 | 0.33 | 0.06 | 0.25 |
| *F03 (°)* | *H* | 0.06 | 0.58 | 0.13 | 0.45 | 0.07 | 0.52 | 0.09 | 0.48 | 0.09 | 0.21 | 0.06 | 0.26 |
| | *V* | <0.05 | 0.71 | <0.05 | 0.69 | <0.05 | 0.72 | <0.05 | 0.70 | 0.14 | 0.24 | <0.05 | 0.64 |
| *F04 (°)* | *H* | 0.99 | 0.79 | 0.41 | 0.73 | 0.99 | 0.71 | 0.10 | 0.72 | 0.53 | 0.46 | <0.05 | 0.70 |
| | *V* | 0.13 | 0.74 | 0.06 | 0.70 | 0.24 | 0.58 | 0.78 | 0.68 | 0.16 | 0.08 | <0.05 | 0.52 |
| *F05 (°)* | *H* | <0.05 | 0.95 | <0.05 | 0.96 | <0.05 | 0.87 | <0.05 | 0.90 | 0.13 | 0.30 | <0.05 | 0.60 |
| | *V* | <0.05 | 0.94 | <0.05 | 0.96 | <0.05 | 0.87 | <0.05 | 0.90 | <0.05 | 0.65 | <0.05 | 0.62 |
| *F06 (°/s)* | *H* | 0.54 | 0.78 | 0.55 | 0.74 | 0.15 | 0.69 | 0.32 | 0.71 | <0.05 | 0.62 | <0.05 | 0.59 |
| | *V* | 0.17 | 0.73 | <0.05 | 0.83 | <0.05 | 0.80 | 0.33 | 0.69 | 0.34 | 0.21 | <0.05 | 0.58 |
| *F07 (°/s)* | *H* | <0.05 | 0.87 | <0.05 | 0.88 | 0.14 | 0.73 | 0.17 | 0.78 | <0.05 | 0.63 | 0.67 | 0.25 |
| | *V* | 0.05 | 0.81 | 0.48 | 0.81 | 0.29 | 0.65 | 0.20 | 0.68 | <0.05 | 0.60 | <0.05 | 0.60 |
| *F08* | *H* | 0.37 | 0.84 | 0.31 | 0.80 | <0.05 | 0.83 | <0.05 | 0.75 | 0.18 | 0.81 | <0.05 | 0.74 |
| | *V* | 0.15 | 0.79 | 0.92 | 0.75 | <0.05 | 0.79 | 0.60 | 0.49 | 0.96 | 0.74 | <0.05 | 0.68 |
| *F09* | *H* | 0.31 | 0.88 | 0.73 | 0.87 | <0.05 | 0.63 | 0.11 | 0.49 | 0.07 | 0.84 | <0.05 | 0.82 |
| | *V* | 0.35 | 0.83 | 0.75 | 0.82 | 0.33 | 0.34 | 0.84 | 0.31 | 0.64 | 0.79 | <0.05 | 0.73 |
| *F14 (°/s)* | *H* | <0.05 | 0.94 | 0.07 | 0.91 | 0.35 | 0.60 | 0.33 | 0.67 | 0.82 | 0.36 | <0.05 | 0.63 |
| | *V* | 0.10 | 0.88 | 0.05 | 0.91 | <0.05 | 0.81 | <0.05 | 0.85 | 0.09 | 0.37 | <0.05 | 0.67 |
| *F15 (°/s)* | *H* | 0.05 | 0.89 | <0.05 | 0.96 | 0.07 | 0.61 | <0.05 | 0.84 | 0.63 | 0.35 | <0.05 | 0.65 |
| | *V* | 0.18 | 0.89 | 0.06 | 0.92 | <0.05 | 0.82 | <0.05 | 0.86 | 0.34 | 0.34 | <0.05 | 0.65 |
| *F16 (°/s)* | *H* | 0.38 | 0.87 | 0.16 | 0.90 | <0.05 | 0.81 | <0.05 | 0.83 | 0.93 | 0.40 | 0.35 | 0.36 |
| | *V* | 0.22 | 0.87 | <0.05 | 0.95 | <0.05 | 0.79 | 0.06 | 0.68 | 0.08 | 0.43 | <0.05 | 0.68 |
| *F17* | *H* | 0.15 | 0.86 | <0.05 | 0.90 | 0.37 | 0.81 | 0.05 | 0.79 | 0.77 | 0.44 | 0.05 | 0.44 |
| | *V* | 0.20 | 0.68 | 0.12 | 0.56 | 0.53 | 0.57 | 0.85 | 0.47 | 0.11 | 0.21 | <0.05 | 0.58 |
| *F18* | *H* | 0.49 | 0.84 | <0.05 | 0.88 | 0.31 | 0.75 | 0.06 | 0.79 | 0.47 | 0.32 | 0.52 | 0.31 |
| | *V* | <0.05 | 0.84 | 0.28 | 0.51 | 0.17 | 0.51 | 0.88 | 0.51 | <0.05 | 0.60 | <0.05 | 0.57 |
| *F19 (%)* | | 0.10 | 0.84 | <0.05 | 0.91 | <0.05 | 0.78 | 0.41 | 0.70 | 0.28 | 0.47 | <0.05 | 0.69 |
| *F20 (%)* | | 0.06 | 0.86 | <0.05 | 0.92 | <0.05 | 0.82 | 0.08 | 0.75 | 0.10 | 0.62 | <0.05 | 0.79 |
| *F21* (°/$s^2$) | *H* | <0.05 | 0.96 | <0.05 | 0.97 | <0.05 | 0.85 | <0.05 | 0.91 | <0.05 | 0.73 | <0.05 | 0.70 |
| | *V* | <0.05 | 0.95 | <0.05 | 0.97 | <0.05 | 0.83 | <0.05 | 0.88 | <0.05 | 0.73 | <0.05 | 0.73 |
| *F22* (°/$s^2$) | *H* | <0.05 | 0.96 | <0.05 | 0.98 | <0.05 | 0.86 | <0.05 | 0.88 | <0.05 | 0.71 | <0.05 | 0.68 |
| | *V* | <0.05 | 0.95 | <0.05 | 0.97 | <0.05 | 0.85 | <0.05 | 0.89 | <0.05 | 0.69 | <0.05 | 0.67 |
| *F23* (°/$s^2$) | *H* | <0.05 | 0.95 | <0.05 | 0.97 | <0.05 | 0.86 | <0.05 | 0.91 | <0.05 | 0.77 | <0.05 | 0.73 |
| | *V* | <0.05 | 0.94 | <0.05 | 0.97 | <0.05 | 0.82 | <0.05 | 0.89 | <0.05 | 0.73 | <0.05 | 0.72 |
| *F24* | *H* | <0.05 | 0.83 | 0.20 | 0.44 | <0.05 | 0.83 | 0.87 | 0.48 | <0.05 | 0.72 | <0.05 | 0.67 |
| | *V* | 0.06 | 0.40 | 0.46 | 0.27 | <0.05 | 0.67 | 0.39 | 0.27 | <0.05 | 0.59 | <0.05 | 0.58 |
| *F25* | *H* | <0.05 | 0.86 | 0.13 | 0.56 | <0.05 | 0.82 | 0.39 | 0.52 | <0.05 | 0.67 | <0.05 | 0.62 |
| | *V* | <0.05 | 0.70 | 0.06 | 0.33 | <0.05 | 0.61 | 0.30 | 0.12 | 0.06 | 0.06 | <0.05 | 0.53 |

In Table 4.2.3 we present the values for the features extracted from saccades. The median duration (*S02*) over the experimental population was calculated to be about 28 ms. This duration seems to be justified given the relatively small amplitude of the saccades performed during reading, and it is within the range reported in other studies that employed the reading paradigm (Abrams, Meyer, & Kornblum, 1989; Nyström & Holmqvist, 2010). The median rate of saccades (*S01*) is similar but slightly lower than the rate of fixations, possibly due to the post-filtering of large saccadic events. A very interesting group of features are those that model the curvature of saccadic shape. As explained, the feature of saccade efficiency (*S05*) models the difference between the amplitude and the actual travelled distance during a saccade. The smaller values of saccade tail efficiency (*S06*) (efficiency at the ending part of saccade) when compared to overall saccade efficiency *S05* can imply the appearance of 'hooks' in saccade shape towards the ending part (when the post-saccadic oscillation phase begins). Qualitative observations of such phenomena have been reported in previous studies (Bahill & Stark, 1975a). The calculated value for the point of maximum raw deviation (*S11*) shows that in general the maximum raw deviation can be expected to occur around the middle (54%) of saccadic trajectory. Since the horizontal component of eye movement is typically more active during the reading task, the values for the dynamic features are much larger than for the vertical component. The median horizontal peak velocity (*S21*) was calculated to be about 170°/s, and the relatively large values of the ***SD***, ***IQ*** feature subtypes reveal a considerable variability of the peak velocity during the duration of a recording. The values of peak acceleration and deceleration (*S27*, *S28*) are both close to 13000°/s$^2$. Similar values but for much smaller population are reported in (Abrams et al., 1989). The median peak acceleration appears to be in overall slightly larger than the peak deceleration, however, the reported variability does not allow to support the generality of this phenomenon. The calculated values for the features of acceleration-deceleration duration ratio (*S39*) and peak acceleration-peak deceleration ratio (*S40*) also suggest the volatility of this difference. The median acceleration-deceleration duration ratio seems to be slightly over one although it is expected that the larger values of peak acceleration (compared to peak deceleration) should correspond to smaller values of duration. An explanation for this discrepancy is that, in general, there is greater difficulty to accurately estimate the exact durations of the acceleration-deceleration phases (non-typical profiles, more than one zero-crossings etc.) compared to the estimation of peak values. The overview of the features of saccadic reading behavior further clarifies the previously discussed difference in fixation and saccade rates (features *F01*, *S01*). In specific, by adding the rates of ‘large’ saccades (*S49*, *S50*) and ‘small’ saccades (*S47*, *S48*) we get a value that is much closer to the fixation rate. The rate of leftward large saccades (*S50*) is 0.4 (about one such saccade per two seconds), and seems to be consisted with the expected rate of line changes during normal reading. The calculated value for the rate of leftward small saccades (*S48*) is 0.8 (about one such saccade per second), a value that seems to be quite large to represent only word regressions. This large value can be attributed to small corrective saccades performed during reading, as for example in order to correct undershoots during line changes (Rayner, 1998).

The overview of the results from assessing the normality and reliability of saccade features is provided in Table 4.2.4. An initial observation is that the percentage of saccade features that are normal (or can be normalized) is much larger (74%) than previously. A prominent grouping of non-normal features seems to occur at some of the kurtosis (***KU***) and skewness (***SK***) feature subtypes. A considerable grouping of non-normal features can be also observed in feature categories *S05*-*S07* (saccade efficiency, tail efficiency, tail inconsistency). For the case of saccade features, the calculated ICC values range from 0.00 to 0.95, with relatively larger percentage (42.1%) of them considered to be highly reliable ('excellent' reliability), 19.9% are considered of 'good' reliability, 16.9% present 'fair' reliability and 21.1% present 'poor' reliability. The feature categories with the top values of ICC are *S36* (the ratio of saccade peak velocity to saccade duration), *S29*-*S31* (modeling of saccade acceleration profile with mean, median, and standard deviation), and *S06* (saccade tail efficiency). All top values refer to horizontal (or radial) components since they are more reliable than vertical components. There are also several other feature categories with exceptional reliability (ICC > 0.9), as for example *S02* (saccade duration) and *S27*-*S28* (peak acceleration and peak deceleration). As previously, the calculated Kendall's W values for the non-normal features seem to be compressed at the upper half of range since they vary from 0.50 to 0.97. The top reliability of feature *S36* (ratio of saccade peak velocity to saccade duration) is further solidified by the higher Kendall's W measure calculated for the ***MN*** subtype (distributional statistic) of this feature, which was designated as non-normal (the rest subtypes were designated normal). The same holds for features *S06* (saccade tail efficiency) and *S02* (saccade duration) when referring to subtypes ***MD***. These and other similar cases (where some subtypes of the same feature category are designated as normal and some as non-normal) seem to imply that although as we mentioned the values of the ICC and Kendall's W measures cannot be directly compared, there is a certain degree of correspondence in their relative assessments about which feature categories are more reliable than the others. Finally, another saccade feature category belonging in the group of non-normal features and showing very high reliability is *S20* (number of local minima in velocity profile).

**Table 4.2.3** Statistics of central tendency and variability for saccade features over the experimental population.

**Non-Distributional Features**

| Feature Type | MD | IQ | Feature Type | MD | IQ | Feature Type | MD | IQ | Feature Type | MD | IQ |
|---|---|---|---|---|---|---|---|---|---|---|---|
| *S01 ($s^{-1}$)* | 3.19 | 0.57 | *S44 (°/s)* | 1.99 | 0.06 | *S48 ($s^{-1}$)* | 0.76 | 0.46 | *S52* | 8.08 | 2.77 |
| *S41 (°)* | -1.45 | 0.85 | *S45 ($s^{-1}$)* | 0.62 | 0.12 | *S49 ($s^{-1}$)* | 0.02 | 0.03 | | | |
| *S42 (°/s)* | 0.14 | 0.05 | *S46* | 0.87 | 0.07 | *S50 ($s^{-1}$)* | 0.40 | 0.13 | | | |
| *S43* | 0.60 | 0.17 | *S47 ($s^{-1}$)* | 2.36 | 0.50 | *S51* | 0.32 | 0.23 | | | |

**Distributional Feature Subtypes**
**Descriptive Statistic: $DistrStat(\cdot)$**

| Feature Type | | *MN* | | *MD* | | *SD* | | *IQ* | | *SK** | | *KU** | |
|---|---|---|---|---|---|---|---|---|---|---|---|---|---|
| | | MD | IQ | MD | IQ | MD | IQ | MD | IQ | MD | IQ | MD | IQ |
| *S02 (ms)* | | 28.01 | 4.87 | 27.00 | 5.00 | 6.79 | 2.57 | 9.00 | 5.00 | 0.42 | 0.64 | 2.97 | 1.41 |
| *S03 (°)* | *H* | 2.46 | 0.50 | 2.26 | 0.53 | 1.29 | 0.35 | 1.66 | 0.54 | 0.91 | 0.49 | 3.86 | 1.60 |
| | *V* | 0.21 | 0.09 | 0.12 | 0.05 | 0.41 | 0.26 | 0.17 | 0.07 | 5.89 | 3.31 | 45.14 | 46.10 |
| *S04 (°)* | | 2.58 | 0.53 | 2.36 | 0.56 | 1.32 | 0.33 | 1.66 | 0.57 | 1.02 | 0.52 | 4.20 | 1.96 |
| *S05* | | 0.97 | 0.01 | 0.98 | 0.01 | 0.03 | 0.02 | 0.02 | 0.01 | -3.03 | 2.05 | 15.13 | 19.05 |
| *S06* | | 0.54 | 0.15 | 0.51 | 0.17 | 0.23 | 0.04 | 0.32 | 0.14 | 0.21 | 0.59 | 2.40 | 0.90 |
| *S07 (%)* | | 48.14 | 10.97 | 42.86 | 14.29 | 22.47 | 6.50 | 23.21 | 21.43 | -0.11 | 0.51 | 3.21 | 1.64 |
| *S08 (°)* | | 1.81 | 6.95 | 1.74 | 5.92 | 38.34 | 4.65 | 41.86 | 10.96 | -0.06 | 0.21 | 2.97 | 0.57 |
| *S09 (°)* | | 5.1E-04 | 2.6E-03 | 6.5E-04 | 2.8E-03 | 0.01 | 0.01 | 0.02 | 0.01 | 0.03 | 0.48 | 3.69 | 1.48 |
| *S10 (°)* | | 1.29 | 2.17 | 2.20 | 2.54 | 5.22 | 1.77 | 6.84 | 2.38 | 0.02 | 0.77 | 4.10 | 3.41 |
| *S11 (%)* | | 53.97 | 5.60 | 56.11 | 8.24 | 27.09 | 2.94 | 43.44 | 9.25 | -0.19 | 0.29 | 2.01 | 0.32 |
| *S12 (°$^2$)* | | 1.68 | 3.18 | 1.42 | 2.76 | 6.78 | 2.95 | 6.75 | 2.53 | 0.22 | 1.55 | 7.97 | 9.94 |
| *S13 (°)* | | 0.02 | 0.04 | 0.02 | 0.04 | 0.09 | 0.04 | 0.09 | 0.04 | 0.21 | 1.40 | 7.31 | 7.97 |
| *S14 (°)* | | 2.64 | 1.41 | 1.82 | 1.51 | 2.87 | 1.11 | 3.34 | 1.49 | 1.75 | 1.03 | 6.83 | 5.51 |
| *S15 (%)* | | 40.95 | 8.42 | 41.65 | 13.51 | 24.86 | 4.89 | 37.77 | 14.55 | -0.15 | 0.49 | 2.13 | 0.70 |
| *S16 (°)* | | -1.42 | 0.88 | -0.55 | 0.89 | 2.09 | 1.00 | 2.03 | 1.37 | -2.50 | 1.51 | 10.45 | 11.13 |
| *S17 (%)* | | 33.37 | 11.60 | 28.13 | 21.69 | 29.55 | 3.73 | 56.96 | 20.87 | 0.29 | 0.67 | 1.77 | 0.51 |
| *S18 (°)* | | 1.25 | 1.94 | 1.61 | 2.11 | 4.37 | 1.53 | 5.29 | 2.10 | 0.09 | 0.84 | 4.55 | 3.10 |
| *S19 (%)* | | 50.48 | 3.18 | 51.42 | 5.11 | 16.16 | 1.43 | 26.51 | 3.66 | -0.21 | 0.28 | 2.27 | 0.50 |
| *S20* | | 0.35 | 0.41 | 0.00 | 0.00 | 0.61 | 0.31 | 1.00 | 1.00 | 1.69 | 1.30 | 5.63 | 5.63 |
| *S21 (°/s)* | *H* | 167.65 | 45.16 | 164.43 | 46.74 | 58.27 | 19.56 | 75.42 | 31.59 | 0.43 | 0.57 | 3.17 | 1.23 |
| | *V* | 27.47 | 9.39 | 21.93 | 7.34 | 24.76 | 12.24 | 14.00 | 6.54 | 4.56 | 2.72 | 30.07 | 32.78 |
| *S22 (°/s)* | *H* | 81.39 | 20.60 | 78.94 | 22.15 | 29.81 | 9.35 | 38.95 | 14.89 | 0.54 | 0.49 | 3.18 | 1.10 |
| | *V* | 11.17 | 3.94 | 8.79 | 3.15 | 10.83 | 5.71 | 5.45 | 2.77 | 5.03 | 2.90 | 34.86 | 37.18 |
| *S23 (°/s)* | *H* | 79.94 | 20.65 | 75.38 | 21.20 | 34.63 | 10.15 | 44.98 | 15.77 | 0.69 | 0.45 | 3.38 | 1.24 |
| | *V* | 9.58 | 3.37 | 7.30 | 2.54 | 10.15 | 6.11 | 5.32 | 2.54 | 5.19 | 3.36 | 36.93 | 42.19 |
| *S24 (°/s)* | *H* | 57.37 | 17.31 | 56.05 | 18.38 | 20.12 | 7.02 | 26.03 | 11.74 | 0.41 | 0.62 | 3.14 | 1.26 |
| | *V* | 8.23 | 2.96 | 6.42 | 2.31 | 8.18 | 3.96 | 4.29 | 2.26 | 4.71 | 2.65 | 31.54 | 32.64 |
| *S25* | *H* | 0.1 | 0.10 | 0.1 | 0.10 | 0.29 | 0.06 | 0.38 | 0.10 | 0.03 | 0.38 | 3.15 | 0.85 |
| | *V* | 0.54 | 0.10 | 0.53 | 0.10 | 0.47 | 0.04 | 0.64 | 0.08 | 0.03 | 0.28 | 2.95 | 0.60 |
| *S26* | *H* | 1.67 | 0.11 | 1.61 | 0.10 | 0.23 | 0.08 | 0.25 | 0.11 | 1.53 | 0.98 | 6.39 | 5.83 |
| | *V* | 2.40 | 0.17 | 2.22 | 0.15 | 0.76 | 0.15 | 0.88 | 0.19 | 1.44 | 0.66 | 5.53 | 3.84 |
| *S27 (°/s$^2$)* | *H* | 13811 | 3519.9 | 13578 | 3702.2 | 4314.1 | 1294.2 | 5659.5 | 1827.2 | 0.27 | 0.54 | 3.00 | 1.04 |
| | *V* | 4694.7 | 1583.6 | 4079.1 | 1315.6 | 2640.2 | 1333.4 | 2660.6 | 922.30 | 2.07 | 1.95 | 9.97 | 13.88 |
| *S28 (°/s$^2$)* | *H* | 12896 | 4369.7 | 12464 | 4281.7 | 4016.3 | 1546.2 | 5161.9 | 1890.6 | 0.50 | 0.56 | 3.28 | 1.38 |
| | *V* | 4929.0 | 1633.4 | 4445.0 | 1430.7 | 2613.6 | 1313.4 | 2755.5 | 969.99 | 1.74 | 2.04 | 8.52 | 13.77 |
| *S29 (°/s$^2$)* | *H* | 6245.2 | 2386.7 | 6122.7 | 2364.0 | 1703.5 | 818.87 | 2150.0 | 1082.9 | 0.35 | 0.56 | 3.37 | 1.13 |
| | *V* | 2075.4 | 651.90 | 1880.0 | 625.15 | 948.74 | 469.65 | 920.06 | 340.21 | 2.25 | 2.24 | 11.25 | 15.48 |
| *S30 (°/s$^2$)* | *H* | 5712.7 | 2463.4 | 5452.6 | 2448.6 | 1916.6 | 907.11 | 2483.1 | 1265.5 | 0.52 | 0.48 | 3.24 | 1.12 |
| | *V* | 1737.4 | 558.01 | 1565.9 | 506.84 | 879.58 | 400.75 | 872.65 | 316.58 | 2.07 | 2.06 | 10.32 | 14.98 |
| *S31 (°/s$^2$)* | *H* | 4197.9 | 1274.2 | 4126.4 | 1254.3 | 1194.3 | 463.70 | 1514.3 | 580.34 | 0.29 | 0.61 | 3.19 | 1.21 |
| | *V* | 1552.0 | 499.58 | 1408.1 | 456.93 | 730.35 | 413.22 | 714.94 | 273.77 | 2.06 | 2.24 | 10.18 | 16.27 |
| *S32* | *H* | 0.43 | 0.25 | 0.43 | 0.23 | 0.42 | 0.05 | 0.56 | 0.08 | 0.04 | 0.31 | 3.07 | 0.60 |
| | *V* | 0.74 | 0.09 | 0.72 | 0.09 | 0.44 | 0.04 | 0.57 | 0.08 | 0.29 | 0.29 | 3.11 | 0.66 |
| *S33* | *H* | 2.38 | 0.29 | 2.22 | 0.23 | 0.68 | 0.26 | 0.74 | 0.27 | 1.49 | 0.54 | 5.74 | 2.86 |
| | *V* | 2.88 | 0.21 | 2.64 | 0.18 | 1.01 | 0.19 | 1.15 | 0.22 | 1.51 | 0.55 | 5.74 | 3.10 |
| *S34 (°/s)* | *H* | 84.37 | 21.83 | 81.94 | 23.65 | 30.77 | 9.76 | 39.98 | 15.56 | 0.49 | 0.47 | 3.22 | 1.07 |
| | *V* | 7.31 | 3.13 | 4.48 | 1.90 | 11.59 | 6.04 | 6.12 | 2.83 | 4.97 | 2.77 | 33.85 | 33.81 |
| *S35 ($s^{-1}$)* | *H* | 76.36 | 13.01 | 74.02 | 14.17 | 19.72 | 9.00 | 23.31 | 8.74 | 1.09 | 2.37 | 5.07 | 15.56 |
| | *V* | 536.97 | 367.86 | 167.49 | 36.95 | 1664.6 | 2560.8 | 204.88 | 74.38 | 7.95 | 4.23 | 74.42 | 72.44 |
| *S36 (°/s$^2$)* | *H* | 6060.8 | 2350.4 | 5888.7 | 2461.3 | 1714.2 | 794.94 | 2225.0 | 1111.6 | 0.37 | 0.61 | 3.37 | 1.09 |
| | *V* | 1004.1 | 377.96 | 802.26 | 305.82 | 796.28 | 448.29 | 532.07 | 244.72 | 3.79 | 2.28 | 22.73 | 22.75 |
| *S37* | *H* | 2.02 | 0.12 | 1.96 | 0.10 | 0.34 | 0.23 | 0.35 | 0.08 | 2.03 | 5.02 | 11.99 | 47.13 |
| | *V* | 14.76 | 10.27 | 4.54 | 1.09 | 46.23 | 72.02 | 5.84 | 2.23 | 7.74 | 4.20 | 69.84 | 72.20 |
| *S38* | | 13.05 | 6.11 | 12.20 | 5.73 | 5.84 | 3.06 | 7.27 | 3.84 | 0.88 | 0.50 | 4.13 | 1.92 |
| *S39* | | 1.01 | 0.20 | 0.91 | 0.20 | 0.46 | 0.11 | 0.54 | 0.13 | 1.28 | 0.74 | 4.98 | 3.15 |
| *S40* | *H* | 1.12 | 0.13 | 1.07 | 0.12 | 0.34 | 0.07 | 0.43 | 0.09 | 0.80 | 0.39 | 3.73 | 1.46 |
| | *V* | 1.07 | 0.12 | 0.94 | 0.10 | 0.62 | 0.15 | 0.65 | 0.12 | 1.86 | 1.07 | 8.10 | 7.34 |

**Table 4.2.4** Statistics of normality and reliability for saccade features over the experimental population.

**Non-Distributional Features**

| Feature Type | p | ICC/W | Feature Type | p | ICC/W | Feature Type | p | ICC/W | Feature Type | p | ICC/W |
|---|---|---|---|---|---|---|---|---|---|---|---|
| *S01* ($s^{-1}$) | 0.96 | 0.80 | *S44* (°/s) | 0.77 | 0.83 | *S48* ($s^{-1}$) | 0.56 | 0.83 | *S52* | 0.08 | 0.55 |
| *S41* (°) | 0.07 | 0.61 | *S45* ($s^{-1}$) | 0.99 | 0.79 | *S49* ($s^{-1}$) | <0.05 | 0.67 | | | |
| *S42* (°/s) | 0.59 | 0.72 | *S46* | <0.05 | 0.86 | *S50* ($s^{-1}$) | 0.08 | 0.62 | | | |
| *S43* | 0.62 | 0.76 | *S47* ($s^{-1}$) | 0.23 | 0.79 | *S51* | 0.47 | 0.83 | | | |

**Distributional Feature Subtypes**
**Descriptive Statistic: $DistrStat(\cdot)$**

| Feature Type | | *MN* | | *MD* | | *SD* | | *IQ* | | *SK** | | *KU** | |
|---|---|---|---|---|---|---|---|---|---|---|---|---|---|
| | | p | ICC/W | p | ICC/W | p | ICC/W | p | ICC/W | p | ICC/W | p | ICC/W |
| *S02* (ms) | | 0.17 | 0.92 | <0.05 | 0.95 | 0.38 | 0.89 | <0.05 | 0.94 | <0.05 | 0.86 | <0.05 | 0.83 |
| *S03* (°) | *H* | 0.07 | 0.76 | 0.22 | 0.74 | 0.09 | 0.74 | 0.34 | 0.73 | 0.12 | 0.40 | 0.19 | 0.43 |
| | *V* | 0.40 | 0.60 | 0.36 | 0.63 | <0.05 | 0.64 | 0.38 | 0.65 | 0.05 | 0.27 | 0.40 | 0.28 |
| *S04* (°) | | 0.12 | 0.77 | 0.08 | 0.75 | 0.14 | 0.73 | 0.63 | 0.72 | 0.30 | 0.36 | 0.48 | 0.45 |
| *S05* | | <0.05 | 0.86 | <0.05 | 0.88 | <0.05 | 0.72 | 0.81 | 0.66 | <0.05 | 0.63 | <0.05 | 0.60 |
| *S06* | | 0.56 | 0.94 | <0.05 | 0.96 | 0.11 | 0.79 | <0.05 | 0.90 | 0.06 | 0.84 | <0.05 | 0.89 |
| *S07* (%) | | <0.05 | 0.92 | <0.05 | 0.86 | 0.26 | 0.84 | <0.05 | 0.89 | <0.05 | 0.80 | <0.05 | 0.93 |
| *S08* (°) | | 0.67 | 0.57 | 0.17 | 0.56 | 0.11 | 0.49 | 0.15 | 0.52 | 0.08 | 0.25 | 0.14 | 0.52 |
| *S09* (°) | | 0.05 | 0.56 | 0.06 | 0.46 | <0.05 | 0.90 | <0.05 | 0.92 | <0.05 | 0.50 | <0.05 | 0.61 |
| *S10* (°) | | 0.15 | 0.79 | 0.80 | 0.74 | 0.34 | 0.48 | 0.17 | 0.72 | <0.05 | 0.57 | 0.15 | 0.28 |
| *S11* (%) | | 0.57 | 0.68 | 0.85 | 0.65 | 0.15 | 0.40 | 0.51 | 0.58 | 0.06 | 0.33 | <0.05 | 0.77 |
| *S12* (°²) | | 0.12 | 0.88 | 0.06 | 0.90 | 0.27 | 0.60 | <0.05 | 0.88 | <0.05 | 0.54 | <0.05 | 0.60 |
| *S13* (°) | | 0.07 | 0.89 | 0.20 | 0.90 | 0.06 | 0.61 | <0.05 | 0.87 | <0.05 | 0.54 | <0.05 | 0.56 |
| *S14* (°) | | 0.37 | 0.87 | 0.54 | 0.88 | 0.08 | 0.64 | 0.70 | 0.77 | <0.05 | 0.66 | <0.05 | 0.61 |
| *S15* (%) | | 0.74 | 0.85 | 0.45 | 0.82 | 0.17 | 0.85 | 0.06 | 0.79 | 0.78 | 0.84 | 0.11 | 0.80 |
| *S16* (°) | | 0.87 | 0.85 | <0.05 | 0.93 | 0.16 | 0.53 | 0.09 | 0.84 | <0.05 | 0.66 | <0.05 | 0.61 |
| *S17* (%) | | 0.94 | 0.87 | 0.88 | 0.87 | 0.23 | 0.78 | <0.05 | 0.85 | 0.66 | 0.89 | <0.05 | 0.88 |
| *S18* (°) | | 0.14 | 0.90 | 0.91 | 0.78 | <0.05 | 0.83 | 0.85 | 0.74 | <0.05 | 0.60 | 0.23 | 0.28 |
| *S19* (%) | | 0.51 | 0.70 | 0.99 | 0.56 | 0.35 | 0.49 | 0.65 | 0.57 | 0.10 | 0.42 | 0.32 | 0.26 |
| *S20* | | <0.05 | 0.95 | <0.05 | 0.94 | 0.11 | 0.87 | <0.05 | 0.88 | 0.06 | 0.84 | <0.05 | 0.84 |
| *S21* (°/s) | *H* | 0.06 | 0.91 | 0.32 | 0.90 | 0.06 | 0.84 | 0.28 | 0.80 | 0.75 | 0.58 | <0.05 | 0.73 |
| | *V* | 0.57 | 0.73 | 0.11 | 0.79 | 0.21 | 0.37 | 0.44 | 0.69 | 0.61 | 0.17 | 0.28 | 0.20 |
| *S22* (°/s) | *H* | 0.43 | 0.90 | 0.15 | 0.88 | 0.22 | 0.86 | 0.69 | 0.79 | 0.12 | 0.49 | 0.20 | 0.45 |
| | *V* | 0.27 | 0.74 | 0.06 | 0.81 | 0.67 | 0.31 | 0.06 | 0.68 | 0.08 | 0.22 | 0.38 | 0.23 |
| *S23* (°/s) | *H* | 0.53 | 0.88 | 0.13 | 0.86 | 0.67 | 0.84 | 0.75 | 0.78 | 0.77 | 0.36 | 0.05 | 0.30 |
| | *V* | 0.14 | 0.71 | 0.08 | 0.81 | 0.17 | 0.30 | 0.08 | 0.69 | 0.51 | 0.25 | 0.17 | 0.24 |
| *S24* (°/s) | *H* | 0.74 | 0.92 | 0.35 | 0.91 | 0.40 | 0.86 | 0.74 | 0.83 | 0.27 | 0.64 | 0.23 | 0.47 |
| | *V* | 0.11 | 0.73 | <0.05 | 0.89 | 0.55 | 0.36 | 0.05 | 0.68 | 0.32 | 0.18 | 0.10 | 0.21 |
| *S25* | *H* | 0.06 | 0.83 | 0.13 | 0.81 | 0.18 | 0.73 | 0.07 | 0.69 | <0.05 | 0.60 | <0.05 | 0.55 |
| | *V* | 0.18 | 0.68 | 0.50 | 0.58 | 0.12 | 0.11 | 0.07 | 0.00 | 0.06 | 0.00 | <0.05 | 0.50 |
| *S26* | *H* | <0.05 | 0.92 | <0.05 | 0.93 | 0.23 | 0.59 | 0.37 | 0.83 | <0.05 | 0.62 | <0.05 | 0.61 |
| | *V* | 0.13 | 0.75 | 0.90 | 0.61 | 0.94 | 0.40 | 0.06 | 0.49 | 0.17 | 0.05 | 0.09 | 0.12 |
| *S27* (°/s²) | *H* | 0.06 | 0.91 | 0.24 | 0.90 | 0.05 | 0.69 | 0.37 | 0.78 | <0.05 | 0.71 | <0.05 | 0.67 |
| | *V* | 0.49 | 0.76 | 0.84 | 0.82 | 0.15 | 0.44 | 0.59 | 0.62 | <0.05 | 0.62 | <0.05 | 0.61 |
| *S28* (°/s²) | *H* | 0.07 | 0.93 | 0.62 | 0.92 | 0.22 | 0.72 | 0.51 | 0.81 | <0.05 | 0.74 | <0.05 | 0.69 |
| | *V* | 0.97 | 0.79 | 0.62 | 0.85 | 0.38 | 0.44 | 0.26 | 0.66 | <0.05 | 0.68 | <0.05 | 0.67 |
| *S29* (°/s²) | *H* | 0.05 | 0.95 | 0.76 | 0.95 | 0.13 | 0.82 | 0.61 | 0.85 | <0.05 | 0.74 | <0.05 | 0.69 |
| | *V* | 0.52 | 0.81 | 0.55 | 0.87 | 0.80 | 0.39 | 0.40 | 0.69 | <0.05 | 0.64 | <0.05 | 0.62 |
| *S30* (°/s²) | *H* | 0.06 | 0.95 | 0.14 | 0.95 | 0.21 | 0.87 | 0.15 | 0.88 | <0.05 | 0.72 | <0.05 | 0.64 |
| | *V* | 0.35 | 0.82 | 0.12 | 0.88 | 0.56 | 0.41 | 0.06 | 0.70 | <0.05 | 0.65 | <0.05 | 0.63 |
| *S31* (°/s²) | *H* | 0.13 | 0.93 | 0.63 | 0.93 | 0.14 | 0.73 | 0.09 | 0.81 | <0.05 | 0.70 | <0.05 | 0.68 |
| | *V* | 0.30 | 0.78 | 0.66 | 0.85 | 0.05 | 0.44 | 0.22 | 0.64 | <0.05 | 0.66 | <0.05 | 0.63 |
| *S32* | *H* | 0.24 | 0.90 | 0.47 | 0.86 | 0.45 | 0.49 | 0.20 | 0.27 | 0.59 | 0.10 | 0.34 | 0.00 |
| | *V* | 0.16 | 0.65 | 0.27 | 0.55 | 0.11 | 0.15 | 0.07 | 0.07 | 0.17 | 0.15 | 0.09 | 0.01 |
| *S33* | *H* | <0.05 | 0.93 | <0.05 | 0.91 | 0.09 | 0.78 | 0.05 | 0.76 | 0.07 | 0.04 | 0.14 | 0.04 |
| | *V* | 0.06 | 0.66 | 0.07 | 0.60 | 0.92 | 0.37 | 0.28 | 0.37 | <0.05 | 0.51 | 0.79 | 0.04 |
| *S34* (°/s) | *H* | 0.50 | 0.90 | 0.22 | 0.89 | 0.13 | 0.86 | 0.27 | 0.80 | 0.08 | 0.49 | 0.24 | 0.45 |
| | *V* | 0.05 | 0.65 | 0.37 | 0.69 | 0.26 | 0.31 | 0.18 | 0.66 | 0.11 | 0.21 | 0.21 | 0.22 |
| *S35* ($s^{-1}$) | *H* | 0.05 | 0.59 | 0.77 | 0.92 | <0.05 | 0.72 | 0.56 | 0.78 | <0.05 | 0.60 | <0.05 | 0.57 |
| | *V* | 0.25 | 0.10 | 0.06 | 0.58 | <0.05 | 0.52 | 0.77 | 0.41 | <0.05 | 0.54 | <0.05 | 0.54 |
| *S36* (°/s²) | *H* | <0.05 | 0.98 | 0.07 | 0.96 | 0.08 | 0.87 | 0.41 | 0.87 | 0.10 | 0.42 | 0.29 | 0.34 |
| | *V* | 0.25 | 0.77 | 0.05 | 0.86 | 0.05 | 0.30 | 0.07 | 0.66 | 0.60 | 0.12 | 0.94 | 0.12 |
| *S37* | *H* | <0.05 | 0.78 | 0.45 | 0.80 | <0.05 | 0.62 | 0.64 | 0.46 | <0.05 | 0.59 | <0.05 | 0.59 |
| | *V* | 0.20 | 0.09 | 0.80 | 0.62 | <0.05 | 0.52 | 0.61 | 0.42 | <0.05 | 0.54 | <0.05 | 0.55 |
| *S38* | | 0.39 | 0.90 | 0.60 | 0.89 | 0.27 | 0.88 | 0.52 | 0.85 | 0.07 | 0.22 | 0.63 | 0.21 |
| *S39* | | 0.23 | 0.87 | <0.05 | 0.95 | 0.26 | 0.60 | 0.60 | 0.61 | 0.11 | 0.38 | 0.27 | 0.24 |
| *S40* | *H* | 0.25 | 0.89 | 0.72 | 0.87 | 0.27 | 0.65 | 0.13 | 0.62 | <0.05 | 0.53 | <0.05 | 0.53 |
| | *V* | 0.14 | 0.60 | 0.43 | 0.56 | 0.77 | 0.31 | 0.51 | 0.36 | 0.06 | 0.18 | <0.05 | 0.55 |

In Table 4.2.5 we show the values for the post-saccadic oscillation features. The median duration (*P01*) of was calculated to be about 14 ms, and the median interval between post-saccadic oscillations (*P02*) was found to be about 400 ms but with high variability. In overall, post-saccadic oscillations seem to occur at more than half of the saccades (*P03*). This finding agrees with previous observations (Nyström & Holmqvist, 2010) and further justifies the necessity for modeling the characteristics of post-saccadic oscillations. The vast majority of post-saccadic oscillations (76.5%) are ‘slow’ (*P04*) (peak velocities between 20°/s and 45°/s), and the percentages of ‘moderate’ (*P05*) (peak velocities between 45°/s and 55°/s) and ‘fast’ (*P06*) (peak velocities larger than 55°/s) post-saccadic oscillations are about 11-12%. The velocity and acceleration profile-modeling features (*P10*-*P20*) demonstrate the intermediate levels of eye mobility compared to saccades and fixations. Also, from the examination of the ratio features we can observe that the post-saccadic oscillations can have 2-3 times smaller duration (*P21*) compared to the preceding saccades, whereas their peak velocities are about 6 times smaller (*P24*) than saccades.

**Table 4.2.5** Statistics of central tendency and variability for post-saccadic oscillation features over the experimental population.

**Non-Distributional Features**

| Feature Type | MD | IQ | Feature Type | MD | IQ |
|---|---|---|---|---|---|
| *P03 (%)* | 61.16 | 35.13 | *P05 (%)* | 10.96 | 7.84 |
| *P04 (%)* | 76.47 | 22.68 | *P06 (%)* | 11.69 | 16.68 |

**Distributional Feature Subtypes**
**Descriptive Statistic: $DistrStat(\cdot)$**

| Feature Type | | *MN* | | *MD* | | *SD* | | *IQ* | | *SK** | | *KU** | |
|---|---|---|---|---|---|---|---|---|---|---|---|---|---|
| | | MD | IQ | MD | IQ | MD | IQ | MD | IQ | MD | IQ | MD | IQ |
| *P01 (ms)* | | 14.04 | 4.63 | 12.00 | 5.00 | 6.83 | 2.50 | 8.00 | 4.63 | 1.09 | 1.34 | 4.15 | 5.73 |
| *P02 (ms)* | | 417.61 | 240.15 | 292.50 | 164.25 | 314.00 | 274.80 | 307.50 | 325.50 | 1.95 | 0.91 | 7.42 | 6.05 |
| *P07 (°)* | *H* | 0.26 | 0.10 | 0.22 | 0.09 | 0.16 | 0.10 | 0.17 | 0.10 | 1.39 | 1.21 | 5.46 | 5.94 |
| | *V* | 0.09 | 0.04 | 0.07 | 0.03 | 0.07 | 0.05 | 0.07 | 0.03 | 1.99 | 2.21 | 8.86 | 14.59 |
| *P08* | *H* | 0.42 | 0.29 | 0.00 | 0.00 | 0.59 | 0.18 | 1.00 | 1.00 | 1.52 | 1.89 | 4.90 | 8.65 |
| | *V* | 0.94 | 0.35 | 1.00 | 0.00 | 0.81 | 0.19 | 1.00 | 0.00 | 0.68 | 0.61 | 3.33 | 1.96 |
| *P09* | *H* | 0.35 | 0.21 | 0.00 | 0.00 | 0.57 | 0.17 | 1.00 | 1.00 | 1.70 | 1.63 | 5.34 | 8.34 |
| | *V* | 0.96 | 0.34 | 1.00 | 0.00 | 0.82 | 0.18 | 1.00 | 0.00 | 0.70 | 0.64 | 3.33 | 1.85 |
| *P10 (°/s)* | *H* | 34.56 | 10.33 | 30.92 | 9.75 | 15.74 | 7.63 | 18.46 | 9.75 | 1.15 | 0.81 | 4.58 | 2.99 |
| | *V* | 15.04 | 5.76 | 13.23 | 5.53 | 8.57 | 4.20 | 9.58 | 3.84 | 1.32 | 1.49 | 5.54 | 7.50 |
| *P11 (°/s)* | *H* | 19.60 | 5.99 | 18.47 | 5.84 | 7.69 | 2.74 | 9.27 | 4.17 | 0.77 | 0.67 | 3.81 | 1.79 |
| | *V* | 7.43 | 3.06 | 6.55 | 2.91 | 4.22 | 1.95 | 4.78 | 2.37 | 1.27 | 1.25 | 5.19 | 6.07 |
| *P12 (°/s)* | *H* | 19.38 | 6.06 | 18.25 | 5.75 | 8.41 | 2.94 | 9.53 | 4.55 | 0.82 | 0.64 | 3.87 | 1.77 |
| | *V* | 6.96 | 3.08 | 5.96 | 2.73 | 4.41 | 2.10 | 5.00 | 2.68 | 1.26 | 1.07 | 5.04 | 4.66 |
| *P13 (°/s)* | *H* | 10.39 | 3.14 | 9.20 | 2.93 | 5.28 | 2.39 | 6.32 | 2.97 | 1.10 | 0.72 | 4.31 | 2.68 |
| | *V* | 4.57 | 1.74 | 3.98 | 1.63 | 2.70 | 1.24 | 3.04 | 1.28 | 1.30 | 1.36 | 5.33 | 7.37 |
| *P14* | *H* | -0.07 | 0.28 | -0.13 | 0.27 | 0.50 | 0.07 | 0.69 | 0.19 | 0.22 | 0.50 | 2.85 | 0.95 |
| | *V* | 0.19 | 0.19 | 0.19 | 0.21 | 0.56 | 0.07 | 0.71 | 0.14 | 0.06 | 0.39 | 3.10 | 0.82 |
| *P15* | *H* | 2.01 | 0.12 | 1.90 | 0.12 | 0.49 | 0.14 | 0.54 | 0.14 | 1.50 | 0.96 | 6.01 | 4.79 |
| | *V* | 2.16 | 0.18 | 1.98 | 0.14 | 0.68 | 0.21 | 0.69 | 0.18 | 1.82 | 0.96 | 7.19 | 5.62 |
| *P16 (°/s$^2$)* | *H* | 2668.1 | 1025.9 | 2550.4 | 959.65 | 994.41 | 465.48 | 1229.7 | 591.04 | 0.72 | 0.62 | 3.59 | 1.72 |
| | *V* | 1590.0 | 717.65 | 1464.5 | 647.84 | 724.89 | 458.57 | 879.24 | 494.30 | 0.91 | 0.87 | 4.15 | 3.36 |
| *P17 (°/s$^2$)* | *H* | 2474.6 | 937.04 | 2341.6 | 900.91 | 1044.8 | 454.21 | 1289.1 | 563.62 | 0.76 | 0.57 | 3.63 | 1.80 |
| | *V* | 1435.5 | 691.71 | 1271.3 | 585.81 | 756.43 | 455.73 | 890.76 | 505.94 | 1.07 | 0.72 | 4.39 | 3.11 |
| *P18 (°/s$^2$)* | *H* | 1722.3 | 683.02 | 1584.8 | 607.61 | 698.26 | 404.35 | 809.86 | 437.07 | 0.92 | 0.80 | 3.98 | 2.63 |
| | *V* | 1089.9 | 472.33 | 992.72 | 417.73 | 510.26 | 297.07 | 608.03 | 300.61 | 0.92 | 1.04 | 4.05 | 3.94 |
| *P19* | *H* | 0.37 | 0.26 | 0.35 | 0.25 | 0.57 | 0.09 | 0.74 | 0.15 | 0.09 | 0.40 | 3.10 | 0.79 |
| | *V* | 0.45 | 0.16 | 0.44 | 0.17 | 0.52 | 0.07 | 0.66 | 0.12 | 0.11 | 0.46 | 3.27 | 0.89 |
| *P20* | *H* | 2.48 | 0.49 | 2.23 | 0.32 | 0.94 | 0.50 | 0.95 | 0.43 | 1.72 | 0.81 | 6.46 | 4.18 |
| | *V* | 2.41 | 0.27 | 2.19 | 0.20 | 0.85 | 0.27 | 0.87 | 0.22 | 1.74 | 0.97 | 6.82 | 5.57 |
| *P21* | | 2.38 | 1.07 | 2.11 | 0.94 | 1.26 | 0.69 | 1.53 | 0.90 | 1.29 | 0.89 | 4.69 | 3.67 |
| *P22 (°/s)* | *H* | 211.93 | 93.61 | 172.35 | 73.29 | 144.95 | 87.20 | 153.23 | 107.83 | 1.56 | 1.06 | 6.06 | 5.57 |
| | *V* | 18.34 | 13.53 | 9.70 | 6.21 | 29.36 | 27.85 | 14.94 | 10.96 | 4.14 | 3.00 | 23.60 | 31.64 |
| *P23* | *H* | 13.59 | 7.74 | 10.12 | 5.84 | 11.35 | 12.36 | 9.97 | 6.60 | 2.72 | 3.06 | 12.69 | 24.22 |
| | *V* | 3.35 | 2.19 | 1.72 | 0.81 | 5.10 | 4.45 | 2.90 | 1.77 | 3.55 | 2.32 | 17.87 | 22.65 |
| *P24* | *H* | 5.76 | 2.24 | 5.02 | 2.00 | 3.14 | 2.02 | 3.34 | 1.59 | 1.82 | 1.80 | 7.46 | 11.25 |
| | *V* | 2.32 | 0.77 | 1.74 | 0.50 | 1.90 | 1.08 | 1.59 | 0.71 | 2.48 | 1.50 | 10.80 | 11.10 |

**Table 4.2.6** Statistics of normality and reliability for post-saccadic oscillation features over the experimental population.

**Non-Distributional Features**

| Feature Type | p | ICC/W | Feature Type | p | ICC/W |
|---|---|---|---|---|---|
| *P03 (%)* | 0.75 | 0.91 | *P05 (%)* | 0.22 | 0.53 |
| *P04 (%)* | 0.05 | 0.83 | *P06 (%)* | 0.54 | 0.80 |

**Distributional Feature Subtypes**
**Descriptive Statistic: $DistrStat(\cdot)$**

| Feature Type | | *MN* | | *MD* | | *SD* | | *IQ* | | *SK* | | *KU* | |
|---|---|---|---|---|---|---|---|---|---|---|---|---|---|
| | | p | ICC/W | p | ICC/W | p | ICC/W | p | ICC/W | p | ICC/W | p | ICC/W |
| *P01 (ms)* | | 0.72 | 0.80 | <0.05 | 0.90 | 0.11 | 0.24 | <0.05 | 0.82 | <0.05 | 0.73 | <0.05 | 0.71 |
| *P02 (ms)* | | <0.05 | 0.94 | <0.05 | 0.94 | 0.59 | 0.82 | <0.05 | 0.94 | <0.05 | 0.62 | <0.05 | 0.63 |
| *P07 (°)* | *H* | 0.26 | 0.83 | 0.20 | 0.85 | <0.05 | 0.83 | <0.05 | 0.87 | <0.05 | 0.67 | <0.05 | 0.64 |
| | *V* | 0.07 | 0.65 | 0.06 | 0.87 | 0.12 | 0.42 | 0.06 | 0.71 | <0.05 | 0.63 | <0.05 | 0.62 |
| *P08* | *H* | 0.80 | 0.61 | <0.05 | 0.88 | <0.05 | 0.73 | <0.05 | 0.85 | <0.05 | 0.82 | <0.05 | 0.75 |
| | *V* | 0.46 | 0.67 | <0.05 | 0.89 | 0.13 | 0.26 | <0.05 | 0.75 | <0.05 | 0.58 | <0.05 | 0.57 |
| *P09* | *H* | 0.11 | 0.49 | <0.05 | 0.93 | <0.05 | 0.72 | <0.05 | 0.87 | <0.05 | 0.77 | <0.05 | 0.72 |
| | *V* | 0.07 | 0.67 | <0.05 | 0.88 | 0.05 | 0.30 | 0.19 | 0.26 | <0.05 | 0.61 | <0.05 | 0.59 |
| *P10 (°/s)* | *H* | <0.05 | 0.93 | <0.05 | 0.92 | 0.06 | 0.72 | 0.06 | 0.82 | <0.05 | 0.68 | 0.06 | 0.33 |
| | *V* | 0.85 | 0.80 | 0.22 | 0.85 | <0.05 | 0.74 | 0.44 | 0.66 | <0.05 | 0.61 | <0.05 | 0.59 |
| *P11 (°/s)* | *H* | 0.84 | 0.87 | 0.05 | 0.87 | 0.05 | 0.78 | 0.44 | 0.79 | <0.05 | 0.70 | 0.77 | 0.31 |
| | *V* | 0.78 | 0.80 | 0.44 | 0.82 | 0.10 | 0.52 | 0.30 | 0.67 | 0.21 | 0.34 | <0.05 | 0.66 |
| *P12 (°/s)* | *H* | 0.27 | 0.84 | 0.30 | 0.82 | 0.90 | 0.73 | <0.05 | 0.87 | 0.23 | 0.30 | 0.56 | 0.24 |
| | *V* | 0.67 | 0.79 | 0.18 | 0.79 | 0.32 | 0.58 | 0.82 | 0.66 | <0.05 | 0.67 | <0.05 | 0.65 |
| *P13 (°/s)* | *H* | <0.05 | 0.92 | <0.05 | 0.90 | 0.06 | 0.74 | <0.05 | 0.87 | 0.06 | 0.42 | 0.64 | 0.35 |
| | *V* | 0.48 | 0.80 | 0.12 | 0.84 | <0.05 | 0.74 | 0.25 | 0.63 | <0.05 | 0.59 | <0.05 | 0.59 |
| *P14* | *H* | 0.49 | 0.79 | <0.05 | 0.86 | 0.09 | 0.24 | 0.37 | 0.41 | 0.05 | 0.42 | 0.17 | 0.35 |
| | *V* | 0.13 | 0.67 | 0.28 | 0.64 | 0.16 | 0.13 | 0.09 | 0.09 | 0.25 | 0.00 | 0.06 | 0.12 |
| *P15* | *H* | 0.06 | 0.45 | 0.13 | 0.58 | 0.05 | 0.21 | 0.16 | 0.31 | <0.05 | 0.57 | <0.05 | 0.57 |
| | *V* | 0.20 | 0.59 | 0.42 | 0.54 | 0.06 | 0.24 | 0.16 | 0.25 | 0.52 | 0.21 | <0.05 | 0.59 |
| *P16 (°/s²)* | *H* | 0.38 | 0.89 | 0.67 | 0.90 | <0.05 | 0.85 | 0.60 | 0.72 | <0.05 | 0.66 | <0.05 | 0.60 |
| | *V* | <0.05 | 0.91 | 0.37 | 0.87 | <0.05 | 0.81 | 0.08 | 0.75 | <0.05 | 0.67 | <0.05 | 0.63 |
| *P17 (°/s²)* | *H* | 0.35 | 0.88 | 0.06 | 0.89 | 0.05 | 0.68 | 0.07 | 0.72 | <0.05 | 0.59 | 0.05 | 0.14 |
| | *V* | <0.05 | 0.90 | 0.23 | 0.85 | <0.05 | 0.82 | 0.05 | 0.75 | <0.05 | 0.65 | 0.15 | 0.29 |
| *P18 (°/s²)* | *H* | 0.94 | 0.89 | 0.65 | 0.92 | <0.05 | 0.84 | 0.06 | 0.79 | <0.05 | 0.68 | <0.05 | 0.62 |
| | *V* | <0.05 | 0.90 | 0.10 | 0.86 | 0.08 | 0.60 | 0.06 | 0.71 | <0.05 | 0.63 | <0.05 | 0.61 |
| *P19* | *H* | 0.77 | 0.75 | 0.09 | 0.71 | 0.33 | 0.49 | 0.22 | 0.34 | 0.26 | 0.01 | 0.84 | 0.03 |
| | *V* | 0.37 | 0.71 | 0.36 | 0.64 | 0.11 | 0.14 | 0.83 | 0.03 | 0.59 | 0.02 | 0.06 | 0.13 |
| *P20* | *H* | 0.29 | 0.83 | <0.05 | 0.90 | <0.05 | 0.84 | 0.05 | 0.66 | 0.27 | 0.22 | <0.05 | 0.57 |
| | *V* | 0.44 | 0.68 | 0.22 | 0.59 | 0.19 | 0.33 | 0.11 | 0.31 | 0.06 | 0.22 | <0.05 | 0.59 |
| *P21* | | <0.05 | 0.93 | <0.05 | 0.92 | <0.05 | 0.88 | 0.05 | 0.79 | <0.05 | 0.72 | <0.05 | 0.68 |
| *P22 (°/s)* | *H* | <0.05 | 0.92 | 0.07 | 0.80 | <0.05 | 0.87 | 0.67 | 0.82 | <0.05 | 0.70 | <0.05 | 0.69 |
| | *V* | <0.05 | 0.83 | <0.05 | 0.86 | 0.07 | 0.41 | <0.05 | 0.87 | 0.51 | 0.28 | 0.07 | 0.33 |
| *P23* | *H* | <0.05 | 0.91 | 0.68 | 0.85 | <0.05 | 0.83 | 0.06 | 0.78 | <0.05 | 0.65 | <0.05 | 0.64 |
| | *V* | <0.05 | 0.83 | 0.16 | 0.68 | <0.05 | 0.76 | <0.05 | 0.82 | <0.05 | 0.62 | <0.05 | 0.62 |
| *P24* | *H* | 0.64 | 0.83 | 0.07 | 0.86 | <0.05 | 0.80 | 0.47 | 0.73 | 0.05 | 0.26 | <0.05 | 0.63 |
| | *V* | 0.36 | 0.73 | 0.75 | 0.69 | <0.05 | 0.77 | <0.05 | 0.80 | <0.05 | 0.58 | <0.05 | 0.58 |

An overview of Table 4.2.6 can reveal the characteristics of normality and reliability of the post-saccadic oscillation features. The percentage of normal (or normalized) post-saccadic oscillation features is slightly higher than for fixations, lying at 55.3%. The ICC for the case of post-saccadic oscillations ranges from 0.00 to 0.92, and the corresponding levels of reliability for the post-saccadic oscillation features are 35.2% ‘excellent’, 23.2% ‘good’, 12.0% ‘fair’, and ‘poor’ 29.6%. Among the most prominent categories of features are *P03* (percentage of saccades followed by a glissade) and *P16*/*P18* (modeling of post-saccadic oscillation acceleration profile with mean and standard deviation). The values of Kendall’s W vary from minimum 0.56 to maximum 0.94, and the most reliable non-normally distributed features appear in categories *P02* (interval between post-saccadic oscillations), *P10* (peak velocity of post-saccadic oscillations), *P21* (ratio of durations of saccades and adjacent post-saccadic oscillations), and *P22* (ratio of amplitudes of saccades and durations of adjacent post-saccadic oscillations).

*4.3. Limitations and Future Extensions*

The current research should be considered within the scope of certain limitations. First of all, the recordings of eye movements were conducted using a specific model of a high-grade eye-tracker (EyeLink 1000). It would be very interesting to investigate the stability of feature values and the results from reliability assessment for eye movement recordings captured with eye-tracker models of different specifications. Since most of the highly reliable features were found to be dynamic and related to velocity/acceleration traces, which are usually the most prone to error, it would be useful to assess the effects of different error sources (eye-tracker dependent etc.) on the features. Second, the current test-retest interval can be considered relatively small (30 min.). Although such an interval justifies a preliminary test-retest analysis for assessing the reliability of features coming from different measurements, the behavior of eye movement features for larger time intervals is expected to shed light on their long-term stability. Last, it is expected that many of the extracted features can be highly correlated, as for example the features from subtypes *MN* and *MD*, or these from *SD* and *IQ*. Although we opted to present all the features for the completeness of the current analysis, in an application scenario, an examination of the pairwise correlations of the features could be possibly performed for the selection of a smaller subset including only the features with the higher reliability (among the highly correlated).

## 5. Conclusion

In this work we presented an extensive overview and analysis of a very large collection of features that can be extracted from eye movements during reading. The described features can be used to model the oculomotor activity during fixations, saccades, and post-saccadic oscillations, and they can be useful during the in-depth examination of the physical and behavioral characteristics of eye movements. Apart from the provided descriptions and methods for the extraction of the features, we also performed an analysis of their typical values, variability, and test-retest reliability using recordings from a large population of subjects. The performed research can provide a useful tool in various studies involving the exploration, analysis, and selection of eye movement features.

## Acknowledgements

This work was supported in part by NSF CAREER grant #CNS-1250718 and NIST grant #60NANB15D325. Special gratitude is expressed to Dr. E. Abdulin and I. S. Vasquez Mondragon for their contribution during the pre-processing of the eye movement recordings.